\documentclass[10pt,twocolumn,letterpaper]{article}

 \usepackage[pagenumbers]{cvpr} 

\usepackage[dvipsnames]{xcolor}

\definecolor{cvprblue}{rgb}{0.21,0.49,0.74}
\usepackage[pagebackref,breaklinks,colorlinks,citecolor=cvprblue]{hyperref}
\usepackage{caption}
\usepackage{pifont}
\usepackage{multirow}
\usepackage{colortbl}
\usepackage{tabularx}
\usepackage{array} 
\usepackage{graphicx}

\definecolor{best}{rgb}{1, 0.5, 0}
\definecolor{second}{rgb}{1, 1, 0}

\usepackage[capitalize]{cleveref}
\crefname{section}{Sec.}{Secs.}
\Crefname{section}{Section}{Sections}
\Crefname{table}{Table}{Tables}
\crefname{table}{Tab.}{Tabs.}

\def\paperID{3501} 
\def\confName{CVPR}
\def\confYear{2025}

\title{CRiM-GS: Continuous Rigid Motion-Aware Gaussian Splatting \\ from Motion-Blurred Images}

\author{{\fontsize{11}{11}\selectfont Jungho Lee$^1$ \quad Donghyeong Kim$^1$ \quad Dogyoon Lee$^1$ \quad Suhwan Cho$^1$ \quad Minhyeok Lee$^1$ \quad Sangyoun Lee$^1$}\\ \\
{\fontsize{11}{11}\selectfont$^1$School of Electrical and Electronic Engineering, Yonsei University}\\
{\tt\small \{2015142131, 2donghyung87, nemotio, chosuhwan, hydragon516, syleee\}@yonsei.ac.kr}\\
{\fontsize{11}{11}\selectfont \tt \href{https://Jho-Yonsei.github.io/CRiM-Gaussian/}{\texttt{https://Jho-Yonsei.github.io/CRiM-Gaussian}}}
}

\begin{document}
\twocolumn[{
	\renewcommand\twocolumn[1][]{#1}
	\maketitle
	\begin{center}
		\centering
		\captionsetup{type=figure}
		\vspace{-3mm}
		\includegraphics[width=1\linewidth]{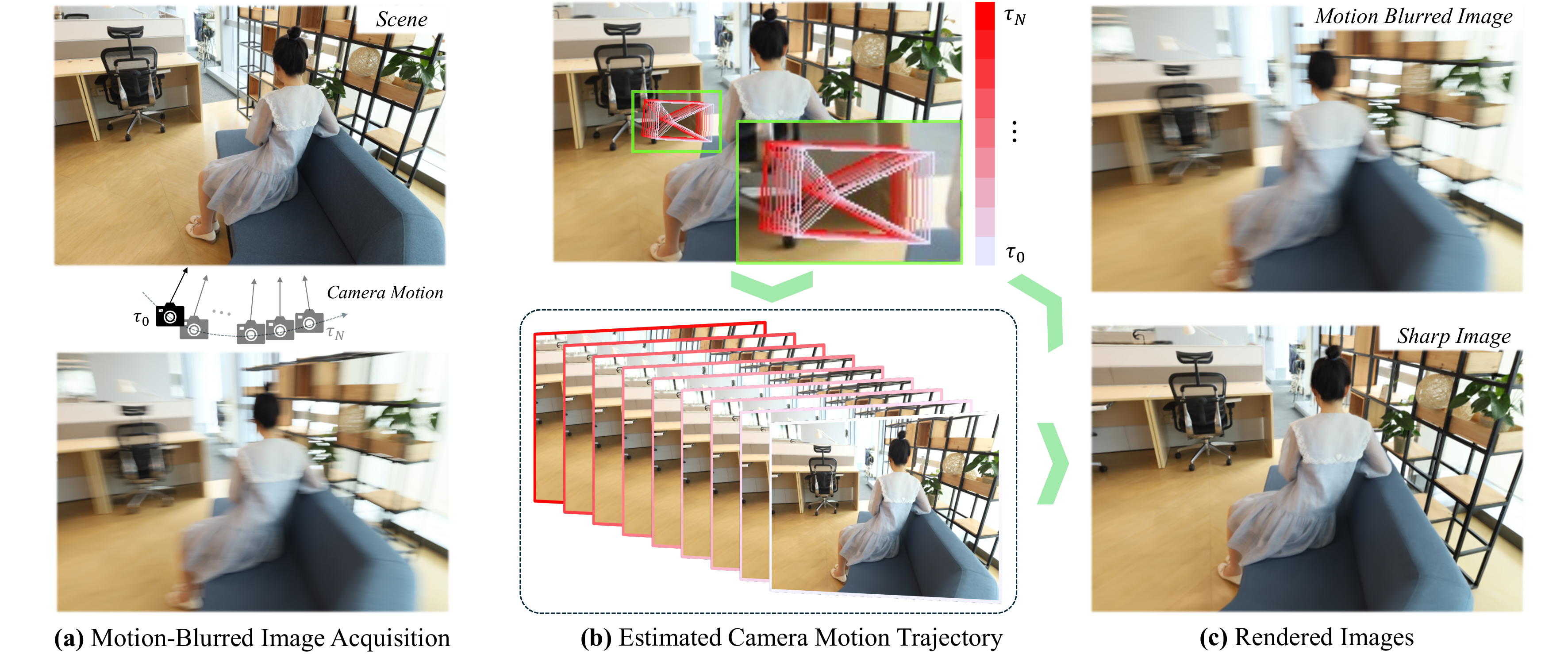}
		\vspace{-6mm}
		\caption{We propose CRiM-GS, a novel framework for reconstructing 3D scenes from camera motion-blurred images. \textbf{(a)} Camera motion-blurred images are generated by the continuous movement of the camera during the exposure time. \textbf{(b)} The top image represents the camera poses obtained through CRiM-GS, showing a continuous trajectory indicated by gradually changing colors. The bottom images are the rendering outputs from each camera pose. \textbf{(c)} By applying a weighted sum to the rendered images obtained in step (b), we get the motion-blurred output image shown at the top, while the sharp output image is rendered from the initial camera pose.}
		\label{fig:teaser}
	\end{center}
}]
\begin{abstract}
3D Gaussian Splatting (3DGS) has gained significant attention for their high-quality novel view rendering, motivating research to address real-world challenges. A critical issue is the camera motion blur caused by movement during exposure, which hinders accurate 3D scene reconstruction. In this study, we propose CRiM-GS, a \textbf{C}ontinuous \textbf{Ri}gid \textbf{M}otion-aware \textbf{G}aussian \textbf{S}platting that reconstructs precise 3D scenes from motion-blurred images while maintaining real-time rendering speed. Considering the complex motion patterns inherent in real-world camera movements, we predict continuous camera trajectories using neural ordinary differential equations (ODE). To ensure accurate modeling, we employ rigid body transformations with proper regularization, preserving object shape and size. Additionally, we introduce an adaptive distortion-aware transformation to compensate for potential nonlinear distortions, such as rolling shutter effects, and unpredictable camera movements. By revisiting fundamental camera theory and leveraging advanced neural training techniques, we achieve precise modeling of continuous camera trajectories. Extensive experiments demonstrate state-of-the-art performance both quantitatively and qualitatively on benchmark datasets..
\end{abstract}
    
\section{Introduction}
\label{sec:intro}

Novel view synthesis has recently garnered significant attention, with neural radiance fields (NeRF)~\cite{mildenhall2020nerf} making considerable advancements in photo-realistic neural rendering. NeRF takes sharp 2D images from multiple views as input to reconstruct precise 3D scenes, which are crucial for applications such as augmented reality (AR) and virtual reality (VR). NeRF models radiance and volume density through an implicit neural representation using multi-layer perceptrons (MLPs), based on the coordinates and viewing directions of 3D points. However, inefficient memory usage of this volume rendering process poses challenges for real-time applications. To address this, 3D Gaussian Splatting (3DGS)~\cite{kerbl20233d} has recently emerged, offering an alternative by explicitly representing 3D scenes and enabling high-quality real-time rendering through a differentiable splatting method.

However, for precise 3D scene representation in real-world scenarios, it is essential to address various forms of image quality degradation, such as camera motion blur and defocus blur. Current methods like NeRF and 3DGS rely on sharp images as input, which assumes highly ideal conditions. Obtaining such precise images requires a large depth of field (DoF), necessitating a very small aperture setting~\cite{hecht2012optics}. However, a smaller aperture limits light intake, leading to longer exposure times. With longer exposure, even slight camera movement introduces complex motion blur in the images. Therefore, developing techniques that can handle motion blur in input images is crucial for accurate 3D scene reconstruction. Our research tackles this challenge by focusing on reconstructing precise 3D scenes from images affected by camera motion blur, thus broadening the applicability of neural rendering in more realistic conditions.

Recently, several methods have been proposed to achieve sharp novel view rendering from camera motion-blurred images. Inspired by traditional blind image deblurring techniques, Deblur-NeRF~\cite{ma2022deblurnerf} firstly introduces a method for deblurring 3D scenes. This approach designs a learnable kernel that intentionally blurs images during training, and during rendering, excludes the learned blurring kernel to render sharp novel view images. Following this approach, various methods targeting camera motion blur~\cite{lee2023dp,lee2024deblurring,wang2023bad,zhao2024bad,peng2023pdrf,peng2024bags} have emerged, aiming to improve blurring kernel estimation accuracy. However, all of these methods predict the camera trajectory without addressing for the continuity of camera motion throughout the exposure time (\textit{e.g}., simple spline function~\cite{wang2023bad,zhao2024bad}). While these approaches effectively model simple motion blur, they tend to oversmooth the trajectory in cases of complex camera motion.

In this paper, we propose a \textbf{C}ontinuous \textbf{Ri}gid \textbf{M}otion-aware \textbf{G}aussian \textbf{S}platting, namly CRiM-GS, a novel approach with three key contributions. First, we apply neural ordinary differential equations (ODE)~\cite{chen2018neural} to model continuous camera movement during exposure time, as illustrated in~\cref{fig:teaser}. By continuously modeling the camera trajectory in 3D space, we introduce a model that is fundamentally different from existing methods~\cite{ma2022deblurnerf,lee2023dp,peng2023pdrf,wang2023bad}. Second, inspired by Nerfies~\cite{park2021nerfies} and DP-NeRF~\cite{lee2023dp}, we incorporate rigid body transformations with explicit regularization, to accurately capture the shape and size of the static subject during camera movement. Third, to address potential nonlinear distortions that can occur during fast camera motion, such as small rolling-shutter effects, we propose an adaptive distortion-aware transformation. This transformation softly refines the rigid body transformation with a higher degree of freedom. For superior rendering quality with real-time rendering speed, we employ Mip-Splatting~\cite{yu2024mip}, a differentiable rasterization-based method based on 3DGS~\cite{kerbl20233d}, rather than ray tracing-based methods (\textit{e.g}., NeRF~\cite{mildenhall2020nerf}, TensoRF~\cite{chen2022tensorf}). We evaluate and compare our approach on Deblur-NeRF~\cite{ma2022deblurnerf} synthetic and real-world scene datasets, achieving state-of-the-art results across the benchmarks. To demonstrate the effectiveness of CRiM-GS, we conduct various ablative experiments on the proposed contributions.
\section{Related Work}
\label{sec:relatedwork}

\subsection{Neural Rendering}
Neural rendering has seen explode in the fields of computer graphics and vision area thanks to the emergence of volumetric rendering and ray tracing-based neural radiance fields~(NeRF)~\cite{mildenhall2020nerf}, which provide realistic rendering quality from 3D scenes. 
NeRF has led to a wide range of studies, including 3D mesh reconstruction~\cite{wang2021neus,wang2022neuris,li2023neuralangelo,sun2021neuralrecon}, dynamic scene~\cite{pumarola2021dnerf,park2021nerfies,park2021hypernerf,li2022neural3dvideo,tretschk2021nonrigid,li2021nsff}, and human avatars~\cite{weng2022humannerf,peng2021animatable,jiang2022neuman}. 
At the same time, several researches have aimed at overcoming slow rendering speed, such as TensoRF~\cite{chen2022tensorf}, Plenoxel~\cite{fridovich2022plenoxels}, and Plenoctree~\cite{yu2021plenoctrees}.
Among these advancements, the recent emergence of 3D Gaussian splatting~(3DGS)~\cite{kerbl20233d}, which offers remarkable performance along with fast training and rendering speed, has further accelerated research in the neural rendering. 
In addition to the aforementioned diverse scenarios for 3D scene modeling, there has been active research focusing on the external non-ideal conditions of given images, such as sparse-view images~\cite{niemeyer2022regnerf,yang2023freenerf,wang2023sparsenerf,wynn2023diffusionerf}, and the absence of camera parameters~\cite{jeong2021self,wang2021nerf,bian2023nope}.
Moreover, there has been significant research attention on non-ideal conditions inherent to the images themselves, such as low-light~\cite{mildenhall2022nerfinthedark,pearl2022nan}, blur~\cite{ma2022deblurnerf,lee2023dp,peng2023pdrf,wang2023bad,lee2024smurf,chen2024deblur,zhao2024bad,peng2024bags}, and inconsistent appearance condition~\cite{martin2021nerfinthewild}. 
Recently, neural rendering from blurred images, which can occur during image acquisition, has attracted attention due to its practical applicability. 

\subsection{Neural Rendering from Blurry Images}
Deblur-NeRF~\cite{ma2022deblurnerf} firstly introduced the deblurred neural radiance fields by importing the blind-deblurring mechanism into the NeRF framework. 
They introduce specific blur kernel in front of the NeRF framework imitating the blind deblurring in 2D image deblurring area.
After the emergence of~\cite{ma2022deblurnerf}, several attempts have been proposed to model the precise blur kernel with various types of neural rendering baseline, such as TensoRF~\cite{chen2022tensorf}, and 3DGS~\cite{kerbl20233d}.
DP-NeRF~\cite{lee2023dp} proposes rigid blur kernel that predict the camera motion during image acquisition process as 3D rigid body motion to preserve the geometric consistency across the scene.
BAD-NeRF~\cite{wang2023bad} and BAD-Gaussians~\cite{zhao2024bad} similarly predict blur kernel as camera motions based on NeRF~\cite{mildenhall2020nerf} and 3DGS~\cite{kerbl20233d}, which assume the simple camera motion and interpolate them between predicted initial and final poses and design simple spline-based methods. 
PDRF~\cite{peng2023pdrf} proposes progressive blur estimation model with hybrid 2-stage efficient rendering scheme that consists of coarse ray and fine voxel renderer.
BAGS~\cite{peng2024bags} proposes CNN-based multi-scale blur-agonostic degaradation kernel with blur masking that indicates the blurred areas.
\section{Preliminary}
\label{sec:preliminary}

\subsection{3D Scene Blind Deblurring}
For conventional image blind deblurring~\cite{whyte2012non,chakrabarti2016neural,srinivasan2017light}, the blurring kernel $h$ is estimated without any supervision. The process to acquire blurry images is achieved by convolving $h$ with sharp images, where the kernels for the convolution take a fixed grid of size around the pixel location $p$. Deblur-NeRF~\cite{ma2022deblurnerf} applies this algorithm to NeRF~\cite{mildenhall2020nerf}, modeling an adaptive sparse kernel for 3D scene representation from blurry images. Deblur-NeRF acquires the blurry pixel color $\mathbf{c}_{blur}$ by warping the original input ray into multiple rays that constitute the blur, then determining the pixel color from the colors obtained from these rays:
\begin{equation} \label{eq:nerf_blur}
	\mathbf{c}_{blur} = \sum_{i=0}^{N} w_{p}^{i}\mathbf{c}_{p}^{i},~w.r.t.~\sum_{i=0}^{N}w_{p}^{i}=1,
\end{equation}
where $N$ and $i$ respectively denote the number of warped rays for convolution and the correspoding index; $w_{p}$ is the corresponding weight at each ray's location, and $\mathbf{c}_{p}$ represents the pixel color of sharp image. Deblur-NeRF enhances learning efficiency by setting the number of warped rays smaller than the kernel size of 2D convolution. In this paper, instead of warping rays, we propose a method to obtain blurry images by applying \cref{eq:nerf_blur} to multiple rendered images.

\subsection{3D Gaussian Splatting}

Unlike ray tracing-based methods~\cite{mildenhall2020nerf,chen2022tensorf,barron2021mipnerf}, 3DGS~\cite{kerbl20233d} is built on a rasterization-based approach with differentiable 3D Gaussians. These 3D Gaussians are initialized from a sparse point cloud obtained via a Structure-from-Motion (SfM)~\cite{schonberger2016pixelwise,shan2008highmotiondeblurring} algorithm and are defined as follows: 
\begin{equation} \label{eq:gaussian_represent}
	\mathbf{G}(\mathbf{x}) = e^{-\frac{1}{2}(\mathbf{x} - \mu)^{\top}\mathbf{\Sigma}^{-1}(\mathbf{x} - \mu)},
\end{equation}
where $\mathbf{x}\in\mathbb{R}^{3}$ is a point on the Gaussian $\mathbf{G}$ centered at the mean vector $\mu\in\mathbb{R}^{3}$ with an covariance matrix $\mathbf{\Sigma}\in\mathbb{R}^{3\times 3}$. The 3D covariance matrix $\mathbf{\Sigma}$ is derived from a learnable scaling vector $\mathbf{s}\in\mathbb{R}^{3}$ and rotation quaternion $\mathbf{q}\in\mathbb{R}^{4}$, from which the scaling matrix $\mathbf{S}\in\mathbb{R}^{3\times 3}$ and rotation matrix $\mathbf{R}\in\mathbb{R}^{3\times 3}$ are obtained and represented as follows: $\mathbf{\Sigma} = \mathbf{R}\mathbf{S}\mathbf{S}^{\top}\mathbf{R}^{\top}$

For differentiable splatting~\cite{yifan2019differentiable}, the Gaussians in the 3D world coordinate system are projected into the 2D camera coordinate system. This projection uses the viewing transformation $\mathbf{W}\in\mathbb{R}^{3\times 3}$ and the Jacobian $\mathbf{J}\in\mathbb{R}^{2\times 3}$ of the affined approximation of the projective transformation to derive the 2D covariance $\mathbf{\Sigma}^{\textrm{2D}}\in\mathbb{R}^{2\times2}$: $\mathbf{\Sigma}^{\textrm{2D}} = \mathbf{J}\mathbf{W}\mathbf{\Sigma}\mathbf{W}^{\top}\mathbf{J}^{\top}$

Each Gaussian includes a set of spherical harmonics (SH) coefficients and an opacity value $\alpha$ to represent view-dependent color $\mathbf{c}$. The pixel color $\mathbf{c}_{p}$ is then obtained by applying alpha blending to $\mathcal{N}$ ordered Gaussians:
\begin{equation} \label{eq:alpha_blending}
	\mathbf{c}_{p}=\sum_{i \in \mathcal{N}} \mathbf{c}_i \alpha_i \prod_{j=1}^{i-1}\left(1-\alpha_j\right).
\end{equation}

In this paper, we continuously model camera poses to project Gaussians $\mathbf{G}$ onto the 2D camera coordinate system over the exposure time and obtain the final blurry image through a rasterization process of the obtained poses and the Gaussians.

\begin{figure*}[t]
	\centering
	\includegraphics[width=\textwidth]{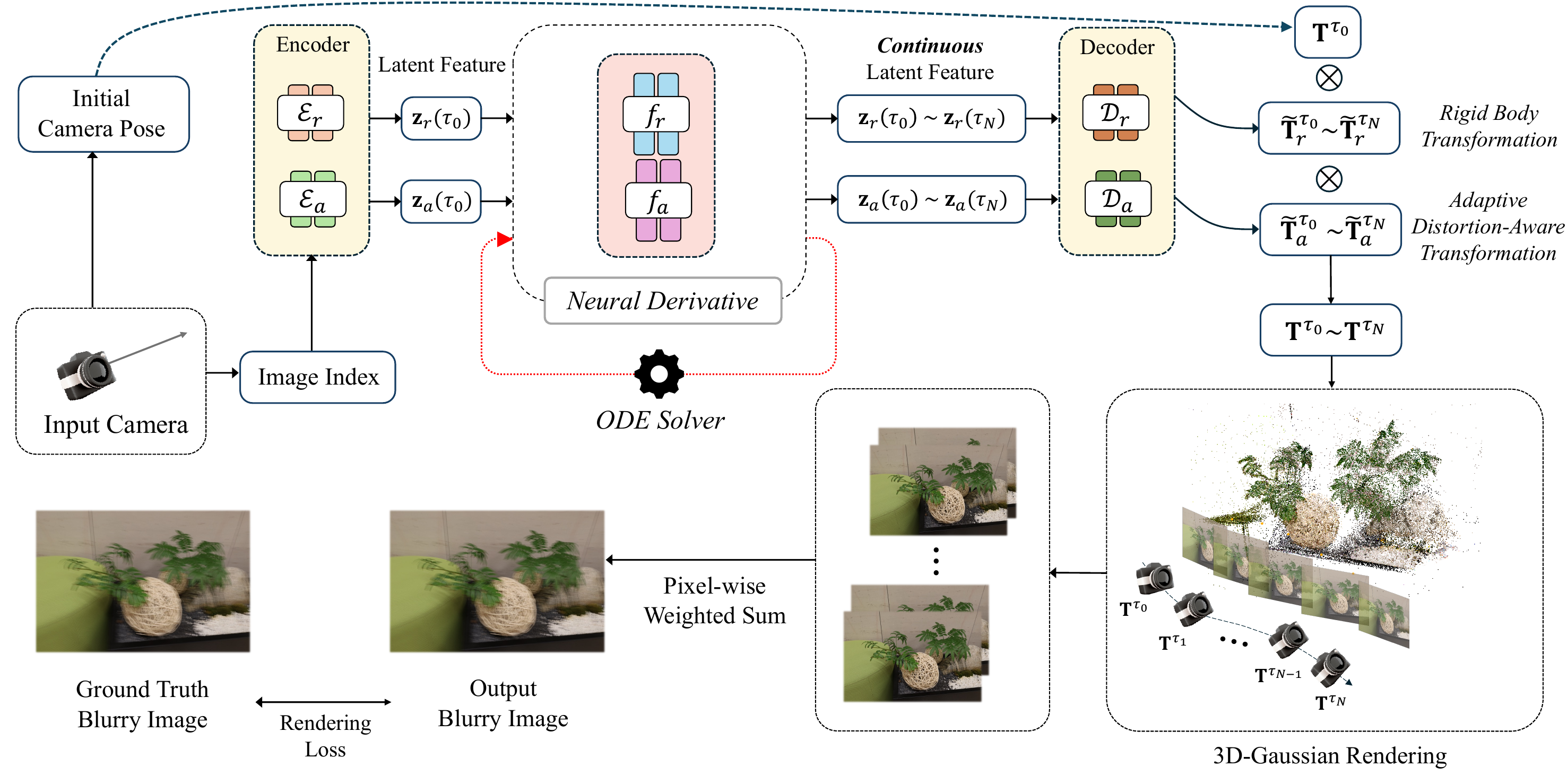}
	\caption{Pipeline of CRiM-GS. The input camera contains the image index and the initial camera pose information. CRiM-GS iteratively solves the ODE using the encoded image index and the neural derivative presented in \cref{sec:rigid} and \cref{sec:deformable}, and obtains $N$ transformed camera poses after decoding. Then, $N$ images are rendered through 3DGS, and a blurry image is obtained through the pixel-wise weighted-sum presented in \cref{sec:optimize}}
	\label{fig:abc}
\end{figure*}

\subsection{Neural Ordinary Differential Equations}

Neural ODE~\cite{chen2018neural} is first proposed as an approach that interprets neural networks as the derivatives of a ODE systems, where ODE represents the dynamics inherent to the hidden states. Specifically, neural ODE are utilized to represent parameterized, time-continuous dynamics in the latent space, providing a unique solution given an initial value and numerical differential equation solvers~\cite{lindelof1894application}. Recently, there has been extensive research on neural ODE. For instance, Latent-ODE~\cite{rubanova2019latent} models the continuous dynamics of irregularly sampled time-series data, while Vid-ODE~\cite{park2021vid} models a smooth latent space by capturing the underlying continuous dynamics of videos.

Neural ODE models a continuous and differentiable latent state $\mathbf{z}(\tau)$. Within an infinitesimally small step limit $\epsilon$ in the latent space, the local continuous dynamics are modeled as $\mathbf{z}(\tau+\epsilon) = \mathbf{z}(\tau) + \epsilon \cdot \frac{d\mathbf{z}(\tau)}{d\tau}$. The derivative of the latent state, $\frac{d\mathbf{z}(\tau)}{d\tau}$, is represented by a neural network $f(\mathbf{z}(\tau), \tau; \phi)$ parameterized by learnable parameters $\phi$. The latent state at any arbitrary time $\tau_s$ is obtained by solving the ODE from the initial time $\tau_0$:
\begin{equation}
	\mathbf{z}\left(\tau_{s}\right)=\mathbf{z}\left(\tau_0\right)+\int_{\tau_0}^{\tau_s} f(\mathbf{z}(\tau), \tau; \phi) d \tau.
\end{equation}
The simplest method to solve ODE is the Euler method~\cite{euler1845institutionum}, which is a fixed-step-size first-order solver. Additionally, the Runge-Kutta~\cite{kutta1901beitrag} methods are preferred as a higher-order solvers as they offer enhanced convergence and stability.

The derivative $f$ modeled by the neural network is expressed as a uniformly Lipschitz continuous non-linear function in $\mathbf{z}$ and $\tau$~\cite{ince1956ordinary}. Therefore, the solution obtained through the solver for any given integration interval $(\tau_{i}, \tau_{j})$ is always unique in the integration of the continuous dynamics. We model the non-linear 3D camera trajectory as time-continuous using neural ODE in the latent space to ensure a continuous representation. In our experiments, we adopt the Dormand-Prince-Shampine fifth-order~\cite{dormand1980family} Runge-Kutta solver for all experiments, following \cite{chen2018neural}.
\section{Method}
\label{sec:method}
\subsection{CRiM-GS Framework}

Our goal is to reconstruct a sharp 3D scene using only camera motion-blurred images as input, thereby obtaining deblurred novel view images. Inspired by image blind deblurring methods, we follow the approach of Deblur-NeRF~\cite{ma2022deblurnerf}, learning a kernel that intentionally blurs images and excluding this kernel during rendering to produce sharp images. As illustrated in \cref{fig:teaser}, our blurring process consists of camera poses along the camera motion trajectory, generated continuously in time order through neural ODE~\cite{chen2018neural}. Each pose is composed of a rigid body transformation $\mathbf{T}_{r}=[\mathbf{R}_{r}|\mathbf{t}_{r}]$ (\cref{sec:rigid}), which maintains the shape and size of the subject, and an adaptive-distortion aware transformation $\mathbf{T}_{a}=[\mathbf{R}_{a}|\mathbf{t}_{a}]$ (\cref{sec:deformable}), which accounts for nonlinear distortions such as small rolling-shutter effect. Rigid body transformation exists in a dense \textit{SE(3)} space, whereas adaptive distortion-aware transformation introduces slight deviations to better capture such distortions while remaining close to \textit{SE(3)}. 
Note that $\mathbf{R} \in \mathbb{R}^{3 \times 3}$ and $\mathbf{t} \in \mathbb{R}^{3}$ represent rotation matrix and translation vector, and $\tilde{\mathbf{R}} \in \mathbb{R}^{3 \times 3}$ and $\tilde{\mathbf{t}} \in \mathbb{R}^{3}$ stand for multiplicative offsets of change of the rotation matrix and translation vector, respectively. Given the two transformation matrices $\tilde{\mathbf{T}}_{r}=[\tilde{\mathbf{R}}_{r}|\tilde{\mathbf{t}}_{r}]$ and $\tilde{\mathbf{T}}_{a}=[\tilde{\mathbf{R}}_{a}|\tilde{\mathbf{t}}_{a}]$, the subsequent pose $\mathbf{T}^{\tau_s}$ in the camera motion is derived as follows:
\begin{equation} \label{eq1}
	\mathbf{T}^{\tau_s} = \mathbf{T}^{\tau_0} \tilde{\mathbf{T}}_{r}^{\tau_s} \tilde{\mathbf{T}}_{a}^{\tau_s},\ where~\mathbf{T} = [\mathbf{R}|\mathbf{t}] = \left[\begin{array}{cc} \mathbf{R} & \mathbf{t} \\ 0 & 1 \end{array}\right],
\end{equation}
where $\mathbf{T}^{\tau_0}=[\mathbf{R}^{\tau_0}|\mathbf{t}^{\tau_0}]$ and  $\mathbf{T}^{\tau_s}=[\mathbf{R}^{\tau_s}|\mathbf{t}^{\tau_s}]$ denote the camera poses at the initial state and the time $\tau_s$. By rendering $N$ images from the obtained $N$ camera poses and computing their pixel-wise weighted sum (\cref{sec:optimize}), we obtain the final blurred image. Our framework is shown in \cref{fig:abc}.

\subsection{Continuous Rigid Body Motion} \label{sec:rigid}

To design the camera motion trajectory within the exposure time, we apply rigid body transformation in that the shape and size of the object remain unchanged. Rigid body transformation requires three components for the unit screw axis $\mathcal{S} = (\hat{\omega}, v)$: the unit rotation axis $\hat{\omega}\in\mathbb{R}^3$, the rotation angle $\theta$ about the axis, and the translation component $v\in\mathbb{R}^{3}$ for translation. Considering that the direction and extent of blur vary for each image in a single scene, we embed the image index of the scene to obtain different embedded features for each image as shown in~\cref{fig:abc}. These features are then passed through an encoder $\mathcal{E}_r$ with a single-layer MLP, transforming them into the latent features $\mathbf{z}_{r}(\tau_0)$, which potentially represent for $\hat{\omega}$, $\theta$, and $v$ of $\mathcal{S}$. Then, we model the continuous latent space for the screw axis using a neural ODE~\cite{chen2018neural} to assign the latent continuity to the screw axis in camera motion trajectory. The neural derivative $f$ of the latent features for the screw axis is expressed as $\frac{d\mathbf{z}_{r}(\tau)}{d\tau} = f(\mathbf{z}_{r}(\tau), \tau; \phi)$, and the latent features at an arbitrary time $\tau_s$ can be obtained by ODE solver, numerically integrating $f$ from $\tau_0$ to $\tau_s$:
\begin{equation} \label{eq5}
	\mathbf{z}_{r}\left(\tau_{s}\right)=\mathbf{z}_{r}\left(\tau_0\right)+\int_{\tau_0}^{\tau_s} f(\mathbf{z}_{r}(\tau), \tau; \phi) d \tau.
\end{equation}
To obtain $N$ poses along the camera trajectory, we uniformly sample $N$ time points within the fixed exposure time and get $N$ latent features by applying the~\cref{eq5} to them. We then transform the latent features obtained by the ODE solver into the unit screw axis $\mathcal{S}$ using a single-layer MLP decoder $\mathcal{D}_{r}$. In Nerfies~\cite{park2021nerfies} and DP-NeRF~\cite{lee2023dp}, the angular velocity $\omega = \hat{\omega} \theta$ is modeled and then decomposed into the unit rotation axis $\hat{\omega}$ and the rotation angle $\theta$, making the two elements dependent on each other. However, since $\theta$ represents the rotation amount about the axis, these two elements should be independent. Therefore, we model $\hat{\omega}$ with normalization and $\theta$ independently through the decoder $\mathcal{D}_{r}$:
\begin{equation}
	\mathcal{D}_{r}\left(\mathbf{z}_{r}(\tau)\right) = (\hat{\omega}^{\tau}, \theta^{\tau}, v^{\tau}), \ where\ ||\hat{\omega}^{\tau}|| = 1.
\end{equation}
According to~\cite{lynch2017modernrobotics}, the screw axis $\mathcal{S}^{\tau}=(\hat{\omega}^{\tau}, v^{\tau})$ is a normalized twist, so the infinitesimal transformation matrix $[\mathcal{S}^{\tau}] \in \mathbb{R}^{4 \times 4}$ is represented as follows:
\begin{equation} \label{eq7}
	[\mathcal{S}^{\tau}] = \left[\begin{array}{cc}
		{[\hat{\omega}^{\tau}]} & v^{\tau} \\
		0 & 0
	\end{array}\right] \in \mathfrak{se}(3),
\end{equation}
where $[\hat{\omega}^{\tau}]\in\mathfrak{so}(3)$ is a $3\times 3$ skew-symmetric matrix of vector $\hat{\omega}^{\tau}$. To derive the infinitesimal transformation matrix $[\mathcal{S}^{\tau}] \theta^{\tau} \in \mathfrak{se}(3)$ on the Lie algebra to the transformation matrix $\mathbf{T}_{r}^{\tau} \in \textit{SE(3)}$ in the Lie group, we use the matrix exponential $e^{[\mathcal{S}^{\tau}] \theta^{\tau}}$, whose rotation matrix and translation vector are $\mathbf{R}^{\tau}_{r}=e^{[\hat{\omega}^{\tau}] \theta^{\tau}}$ and $\mathbf{t}^{\tau}_{r}=G(\theta^{\tau}) v^{\tau}$, respectively. These matrices are expressed as follows using Rodrigues' formula~\cite{rodrigues1816attraction} through Taylor expansion:
\begin{equation}\label{eq:9}
	e^{[\hat{\omega}^{\tau}] \theta^{\tau}} = I + \sin \theta^{\tau} [\hat{\omega}^{\tau}] + (1 - \cos \theta^{\tau}) [\hat{\omega}^{\tau}]^2 \in \textit{SO(3)},
\end{equation} \vspace{-3mm}
\begin{equation}\label{eq:10}
	G(\theta^{\tau}) = I \theta^{\tau} + (1 - \cos \theta^{\tau}) [\hat{\omega}^{\tau}] + (\theta^{\tau} - \sin \theta^{\tau}) [\hat{\omega}^{\tau}]^2.
\end{equation}
Through the whole process, we obtain $N$ continuous rigid body transformation matrices $\mathbf{T}_{r}=e^{[\mathcal{S}] \theta}$, and by multiplying these with the input pose as in \cref{eq1}, we obtain the transformed poses. We show whole derivation process of \cref{eq:9,eq:10} in the \textbf{appendix}, referring the Modern Robotics~\cite{lynch2017modernrobotics}.

\subsection{Adaptive Distortion-Aware Transformation} \label{sec:deformable}

In real-world scenarios, blurry images captured with fast camera motion may exhibit nonlinear distortions, such as small rolling shutter effects, which are present but challenging to detect visually. Consequently, such distortions may not be fully addressed by rigid body transformations alone, suggesting the need for more sophisticated models to accurately handle these complexities. Based on this issue, we propose an adaptive distortion-aware transformation to provide additional refinement to the rigid body transformation. This transformation has higher degrees of freedom from a learning perspective and is simple to implement. The adaptive distortion-aware transformation optimizes the transformation matrix close to \textit{SE(3)} directly without any screw axis components as described in \cref{eq7}. Thus, instead of modeling $\hat{\omega}$ and $v$, this transformation designs a matrix $\tilde{\mathbf{R}}_{a}$ which is regularized close to rotation matrix, and the translation vector $\tilde{\mathbf{t}}_{a}$. This process begins by encoding the image index into the latent state $\mathbf{z}_{a}(\tau_0)$ of $\tilde{\mathbf{R}}_{a}$ and $\tilde{\mathbf{t}}_{a}$ using the encoder $\mathcal{E}_{a}$ with a single-layer MLP, similar to \cref{sec:rigid}. The latent state $\mathbf{z}_{a}(\tau_s)$ at any arbitrary time $\tau_s$ is obtained using the neural derivative $g$ parameterized by $\psi$ and a solver:
\begin{equation}
	\mathbf{z}_{a}(\tau_s) = \mathbf{z}_{a}(\tau_0) + \int_{\tau_0}^{\tau_s} g(\mathbf{z}_{a}(\tau), \tau; \psi) \, d\tau.
\end{equation}
The latent features are then decoded into rotation matrix component $\mathbf{A}^{\tau}_{a}$ and translation vector $\tilde{\mathbf{t}}^{\tau}_{a}$ through a single-layer MLP decoder $\mathcal{D}_{a}$:
\begin{equation}
	\mathcal{D}_{a}(\mathbf{z}_{a}(\tau)) = (\mathbf{A}_{a}^{\tau}, \tilde{\mathbf{t}}^{\tau}) \rightarrow \tilde{\mathbf{T}}_{a}^{\tau} = \left[\begin{array}{cc}
		\tilde{\mathbf{R}}_{a}^{\tau} & \tilde{\mathbf{t}}_{a}^{\tau} \\
		0 & 1
	\end{array}\right],
\end{equation}
where $\tilde{\mathbf{R}}^{\tau}_{a} = \mathbf{A}^{\tau}_{a} + \mathbf{I}$ and $\mathbf{I}$ is the identity matrix to initialize $\tilde{\mathbf{R}}^{\tau}_{a}$ to identity transformation. Since the adaptive distortion-aware transformation is intended to softly refine the rigid body transformation, it should not significantly affect the rigid body transformation during the initial stages of training. Therefore, we initialize the weights of the decoder to approximate $\mathbf{A}_{a}^{\tau}$ to zero matrix using a uniform distribution $\mathcal{U}(-10^{-5}, 10^{-5})$, which means that $\tilde{\mathbf{R}}_{a}^{\tau}$ is initialized close to identity matrix. Furthermore, for modeling the transformation matrix $\tilde{\mathbf{T}}_{a}^{\tau}$ to approximate the main conditions of the \textit{SE(3)} space, we introduce a simple regularization loss. The proposed loss is $\mathcal{L}_{o}$, which ensures the orthogonality condition of rotation matrix: $\mathcal{L}_{o} = \|\tilde{\mathbf{R}}_{a}^{\top} \tilde{\mathbf{R}}_{a} - \mathbf{I}\|_2$. This regularization ensures that the transformation matrix $\tilde{\mathbf{T}}_{a}$ closely lies within the \textit{SE(3)} space. Note that this transformation should not be perfectly in the \textit{SE(3)} space as it aims to compensate the subtle distortions during camera motion. Finally, we apply \cref{eq5} using $\tilde{\mathbf{T}}_{a}^{\tau_s}$ and $\tilde{\mathbf{T}}_{r}^{\tau_s}$ obtained from \cref{sec:rigid} to get the refined transformed camera pose $\mathbf{T}^{\tau_s}$. We discuss the relationships between the adaptive distortion-aware transformation and image distortion from camera motion blur in \cref{subsec:rolling_shutter}.

\subsection{Optimization} \label{sec:optimize}
\paragraph{Pixel-wise Weight.}
Once the $N$ images along the camera motion trajectory are obtained from the $N$ camera poses, we apply a pixel-wise weighted sum to create the blurry image $\mathcal{I}_{blur}$, following previous research~\cite{ma2022deblurnerf,lee2023dp,lee2024smurf,peng2023pdrf}. To satisfy ~\cref{eq:nerf_blur}, we use a shallow CNN $\mathcal{F}$ and a softmax function to compute the pixel-wise weights $\mathcal{P} \in \mathbb{R}^{N \times H \times W \times 3}$ for the resulting images, as follows:
\begin{equation}
	\mathcal{I}_{blur} = \sum_{i=1}^{N}{\mathcal{I}^{\tau_i} \cdot \mathcal{P}^{\tau_i}}, ~ where~\mathcal{P} = \textrm{softmax}(\mathcal{F}(\mathcal{I})),
\end{equation}
where $\mathcal{I}^{\tau_i}$ is the $i$-th image along the camera motion, and $\mathcal{P}^{\tau_i}$ is the pixel-wise weight of $i$-th image. Additionally, we adopt per-pixel scalar mask~\cite{peng2024bags} $\mathcal{M}\in\mathbb{R}^{H\times W\times 3}$ to generate final output blurry image $\mathcal{I}_{out}$ by blending the sharp image $\mathcal{I}^{\tau_0}$ at camera pose $\mathbf{T}^{\tau_0}$ and the blurry image $\mathcal{I}_{blur}$ acquired by pixel-wise weight: $\mathcal{I}_{out}=(\mathbf{1} - \mathcal{M})\cdot\mathcal{I}^{\tau_0} + \mathcal{M}\cdot\mathcal{I}_{blur}$. The scalar mask aims to decide whether each pixel is blurry or not, and a small sparsity constraint is applied to the mask to assign a larger weight on $\mathcal{I}^{\tau_0}$, namely mask sparsity loss~\cite{peng2024bags} $\mathcal{L}_{\mathcal{M}}$, which is a mean value of the scalar mask. This method allows us to obtain the final precise blurry image, which is then optimized against the ground truth blurry image.

\begin{table}[!t] 
	\begin{center}
		\caption{Comparisons on synthetic and real-world scene dataset. ``*'' denotes the results obtained by reproducing the released code. The \colorbox{best!25}{orange} and \colorbox{second!35}{yellow} cells respectively indicate the highest and second-highest value.}
		\resizebox{\columnwidth}{!}{
			\centering
			\setlength{\tabcolsep}{1pt}
			\begin{tabular}{l||c|c|c|c|c|c}
				\toprule 
				
				\multirow{2}{*}{Methods} 			   	& \multicolumn{3}{c|}{~Synthetic Scene~ }  	   & \multicolumn{3}{c}{~Real-World Scene~}  	\\ \cmidrule{2-7}
				&~PSNR$\uparrow$~    &~SSIM$\uparrow$~    &~LPIPS$\downarrow$~ &~PSNR$\uparrow$~    &~SSIM$\uparrow$~    &~LPIPS$\downarrow$~     	\\ \midrule \midrule
				Naive NeRF~\cite{mildenhall2020nerf}                            				 & 23.78    & 0.6807   & 0.3362 		& 22.69       & 0.6347      & 0.3687 	 	\\
				NeRF+MPR~\cite{zamir2021multi}                           					& 25.11    & 0.7476   & 0.2148 			& 23.38       & 0.6655      & 0.3140      	\\ \midrule
				Deblur-NeRF~\cite{ma2022deblurnerf}                          			& 28.77    & 0.8593   & 0.1400 		 & 25.63       & 0.7645      & 0.1820 	\\
				PDRF-10*~\cite{peng2023pdrf}                         					   & 28.86   & \cellcolor{second!35}0.8795   & 0.1139 		 		  & 25.90       & 0.7734      & 0.1825        				\\
				BAD-NeRF*~\cite{wang2023bad}										   & 27.32   & 0.8178   & 0.1127 		 		  & 22.82       & 0.6315      & 0.2887        			\\
				DP-NeRF~\cite{lee2023dp}                        						  & \cellcolor{second!35}29.23    & 0.8674   & 0.1184 		& 25.91   	 & 0.7751      & 0.1602   \\ \midrule
				DeblurGS*~\cite{oh2024deblurgs}									& 20.22		& 0.5454	& 0.1042			& 20.48		& 0.5813		& 0.1186 \\
				BAD-Gaussians*~\cite{zhao2024bad}                          			& 21.12   & 0.5852   & \cellcolor{second!35}0.1007 		 & 20.82       & 0.5917      & 0.1000 	\\
				Deblurring 3DGS~\cite{lee2024deblurring}~                        				& 28.24    & 0.8580   & 0.1051 		 & 26.61       & 0.8224      & 0.1096 	\\ 
				BAGS~\cite{peng2024bags}                          				& 27.34    & 0.8353   & 0.1116 		 & \cellcolor{second!35}26.70       & \cellcolor{second!35}0.8237      & \cellcolor{second!35}0.0956 	\\ \midrule \midrule
				\textbf{CRiM-GS}											                        & \cellcolor{best!25}30.46    & \cellcolor{best!25}0.9095   & \cellcolor{best!25}0.0441 		 & \cellcolor{best!25}27.35       & \cellcolor{best!25}0.8315      & \cellcolor{best!25}0.0633	\\ \bottomrule
			\end{tabular}
		}
		\label{tab:comparison}
	\end{center}
	\vspace{-5mm}
\end{table}

\paragraph{Objective.} We optimize the learning process using the $\mathcal{L}_{1}$ loss and D-SSIM between the generated output blurry image $\mathcal{I}_{out}$ and the ground truth blurry image, similar to 3D-GS~\cite{kerbl20233d}. The $\mathcal{L}_{1}$ loss ensures pixel-wise accuracy, while D-SSIM captures perceptual differences. Additionally, we apply the regularization loss $\mathcal{L}_{o}$ (\cref{sec:deformable}) to regularize the calibration-enhanced transformation and mask sparsity loss $\mathcal{L}_\mathcal{M}$. The final objective $\mathcal{L}$ is defined as follows:
\begin{equation}
	\mathcal{L} = (1 - \lambda_{c})\mathcal{L}_{1} + \lambda_{c}\mathcal{L}_{\rm{D-SSIM}} + \lambda_{o}\mathcal{L}_{o} + \lambda_{\mathcal{M}}\mathcal{L}_{\mathcal{M}},
\end{equation}
where $\lambda_{c}$ is a factor for balancing $\mathcal{L}_{1}$ and $\mathcal{L}_{\rm{D-SSIM}}$, and $\lambda_{o}$ and $\lambda_{\mathcal{M}}$ are factors for $\mathcal{L}_{o}$ and $\mathcal{L}_{\mathcal{M}}$, respectively.
\begin{figure*}[t]
	\centering
	\includegraphics[width=\linewidth]{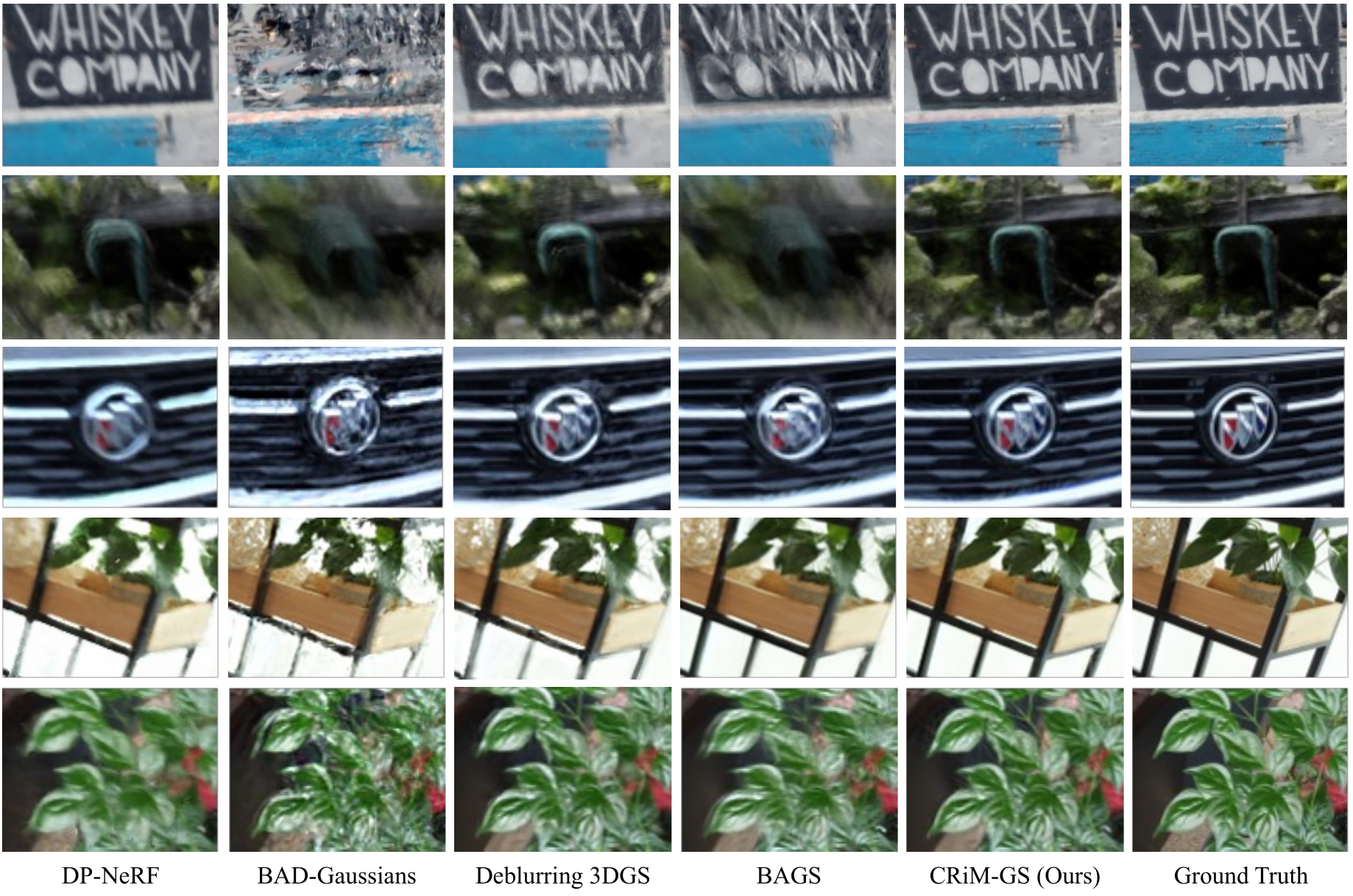}
	\caption{Qualitative comparison on the synthetic and real-world scenes.}
	\label{fig:comparison}
	\vspace{-2mm}
\end{figure*}

\section{Experiments}
\label{sec:experiments}

\paragraph{Datasets.} CRiM-GS is evaluated using the camera motion blur dataset provided by Deblur-NeRF~\cite{ma2022deblurnerf}, which includes 5 synthetic scenes and 10 real-world scenes. The synthetic scenes are generated using Blender~\cite{blender}, and each single image that constitutes a motion-blurred image is obtained through linear interpolation between the first and last camera poses along the camera path. The obtained images are combined in linear RGB space to create the final blurred images. The real-world scenes consist of images captured with a CANON EOS RP camera, where the exposure time is manually set to produce blurry images. For each scene, all images exhibit different types of blur, including non-uniform blur. We obtain the camera poses for each image and the initial point clouds for training the Gaussian primitives by applying COLMAP~\cite{schonberger2016pixelwise,shan2008highmotiondeblurring}. We run COLMAP with blurry images for synthetic and real-world scenes.

\subsection{Novel View Synthesis Results}
For quantitative results, we evaluate CRiM-GS using three metrics: peak signal-to-noise ratio (PSNR), structural similarity index measure (SSIM), and learned perceptual image patch similarity (LPIPS). We compare our method with both ray-based and rasterization methods. Note that the results for BAD-GS~\cite{zhao2024bad} are obtained by re-running the released code ourselves. The overall results for all scenes are shown in \cref{tab:comparison}, demonstrating that our method achieves superior performance compared to all other methods. Specifically for the LPIPS, CRiM-GS shows an improvement of approximately 52\% for synthetic scenes and 33\% for real-world scenes than the state-of-the-art, and the best LPIPS performance across all scenes not only for synthetic dataset but also real-world dataset. Additionally, benefiting from using Mip-Splatting~\cite{yu2024mip} as our backbone, our proposed method ensures fast training and rendering speeds. Although BAD-GS~\cite{zhao2024bad} achieves relatively good LPIPS scores, their PSNR and SSIM scores are lower, indicating that their camera poses are not properly optimized. More details about pose optimization are in the \textbf{appendix}.

\begin{table}[t] 
	\begin{center}
		\caption{Comparisons on synthetic and real-world dataset. We evaluate the performance on three metrics (PSNR, SSIM, LPIPS).}
		\resizebox{\columnwidth}{!}{
			\centering
			\setlength{\tabcolsep}{1pt}
			\begin{tabular}{l||ccc|ccc}
				\toprule 
				\multirow{2}{*}{~Methods~} 		& \multicolumn{3}{c|}{Synthetic Scene}  	   & \multicolumn{3}{c}{Real-World Scene}  	\\ \cmidrule{2-7}
				&													~PSNR$\uparrow$~    &~SSIM$\uparrow$~    &~LPIPS$\downarrow$~ &~PSNR$\uparrow$~    &~SSIM$\uparrow$~    &~LPIPS$\downarrow$~     	\\ \midrule \midrule
				Baseline~\cite{yu2024mip}             						& 21.09    & 0.5974   & 0.2774 		& 21.80       & 0.6135      & 0.2698 	 	\\ \midrule
				CRiM-GS					  							& 		& 	& 			& 		& 		&  \\
				~~~(A) with $\tilde{\mathbf{T}}_{r}$					  														& 26.38		& 0.8355	& 0.0859			& 25.29		& 0.7833		& 0.0985 \\
				~~~(B) with $\tilde{\mathbf{T}}_{r}$ + $\mathcal{P}$					  							& 29.66		& 0.8918	& 0.0546			& 26.77 	& 0.8124		& 0.0778 \\ \midrule				
				~~~(C) with $\tilde{\mathbf{T}}_{a}$					  														& 25.80		& 0.8271	& 0.1052			& 25.09		& 0.7752		& 0.1082 \\
				~~~(D) with $\tilde{\mathbf{T}}_{a}$ + $\mathcal{P}$					  							& 28.79		& 0.8714	& 0.0833			& 26.56		& 0.8070		& 0.0906 \\ \midrule
				~~~(E) with $\tilde{\mathbf{T}}_{r}$ + $\tilde{\mathbf{T}}_{a}$					  				    	& 27.38		& 0.8449	& 0.0762			& 25.52 	& 0.7882		& 0.0973 \\
				~~~(F) with $\tilde{\mathbf{T}}_{r}$ + $\tilde{\mathbf{T}}_{a}$ + $\mathcal{P}$~			& \cellcolor{best!25}30.46    & \cellcolor{best!25}0.9095   & \cellcolor{best!25}0.0441 		 & \cellcolor{best!25}27.35       & \cellcolor{best!25}0.8315      & \cellcolor{best!25}0.0633 \\ \bottomrule
			\end{tabular}
		}
	\end{center}
	\vspace{-4mm}
	\label{tab:ablation}
\end{table}

To qualitatively evaluate the results, we visualize rendering results of several scenes and compare to other methods as shown in \cref{fig:comparison}. These include two synthetic scene and three real-world scenes. Our method shows superior qualitative results compared to the state-of-the-art ray-based method DP-NeRF~\cite{lee2023dp}, as well as rasterization-based methods such as BAD-GS~\cite{zhao2024bad}, BAGS~\cite{peng2024bags}, and Deblurring 3DGS~\cite{lee2024deblurring}. The lowest row of \cref{fig:comparison} stands for visualization of \textsc{Parterre} scene, which shows slightly lower PSNR and SSIM than Deblurring 3DGS, but shows better qualitative performance with well-detailed leafs.

\subsection{Ablation Study}
As shown in \cref{tab:ablation}, to thoroughly demonstrate the validity and effectiveness of our contributions, we conduct ablation experiments on all scenes, not just a single scene.

\paragraph{Ablation on Continuous Transformation.} Referring \cref{tab:ablation}, the model (A) with the main component of CRiM-GS, the continuous rigid body transformation, shows a significant performance increase, indicating that the camera motion path is well-modeled. The model (C) with only the adaptive distortion-aware transformation performs worse than the model with only the rigid body transformation, suggesting that the adaptive distortion-awrare transformation leads to suboptimal results due to its too high degrees of freedom. Therefore, the model (E) combining both transformations, which refines camera motion distortion, achieves the best performance without pixel-wise weights.

\begin{figure}[t]
	\centering
	\includegraphics[width=\linewidth]{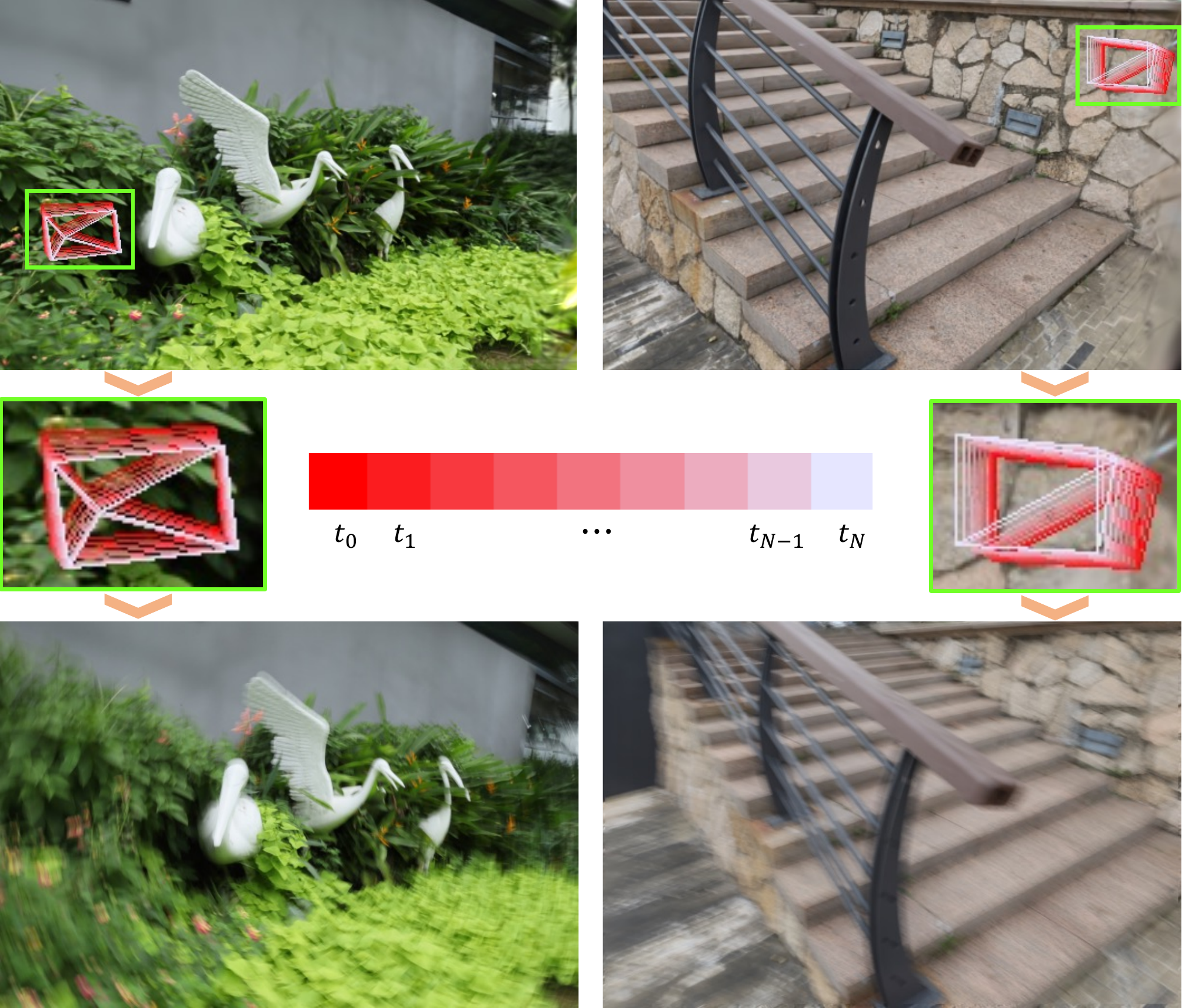}
	\caption{Camera Trajectory Visualization. Red cones stand for camera poses, and the images below are the output blurry images.}
	\label{fig:camera_vis}
	\vspace{-2mm}
\end{figure}

\paragraph{Ablation on Pixel-wise Weight.} The pixel-wise weight, inspired by traditional ray-based 3D scene deblurring methods~\cite{lee2023dp,ma2022deblurnerf,peng2024bags} and blind image deblurring methods~\cite{son2021pvd,zamir2021multi}, differs from the approach of averaging images that constitute a blurry image. As shown in \cref{tab:ablation} for models (B), (D), and (F), applying this method consistently results in higher performance than not applying it. This indicates that by assigning corresponding weights to each pixel location, a more precise blurred image is reconstructed, and the images that constitute the blurry image are modeled more accuratel, surpassing existing state-of-the-art methods.

\subsection{Camera Trajectory Visualization} 
We visualize the continuous camera trajectories for the \textsc{Heron} and \textsc{Stair} scenes from the Deblur-NeRF Real-World dataset, as shown in \cref{fig:camera_vis}. The camera trajectory for a single motion-blurred image is represented by colored cones, with the cone's color gradually transitioning from red to light purple as time progresses from $t_{0}$ to $t_{N}$. The visualized trajectories confirm that the camera paths generated by CRiM-GS are smoothly continuous, validating that the continuous transformations described in \cref{sec:rigid,sec:deformable} function as intended. Additionally, the output blurry images in \cref{fig:camera_vis} align with the visualized camera poses, accurately reflecting both the positions and directions of the camera. Specifically, even though the camera path in the \textsc{Heron} scene on the left is nonlinear, CRiM-GS accurately predicts this path and generates a precise blurry image.

\subsection{Rolling Shutter Effect Compensation}\label{subsec:rolling_shutter}
\begin{figure}[t]
	\centering
	\includegraphics[width=\linewidth]{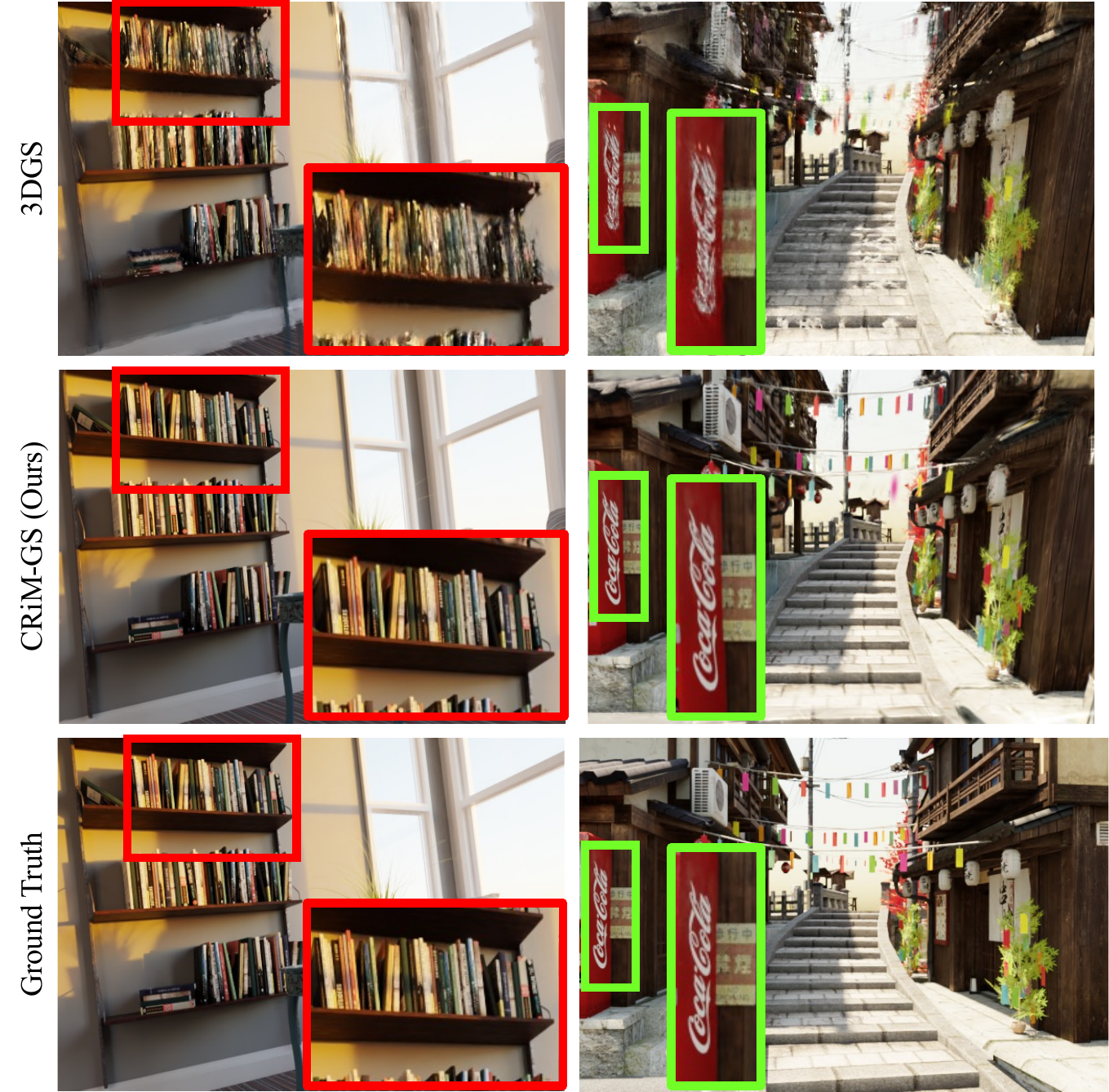}
	\caption{Rendering Results on Rolling Shutter Effect Dataset~\cite{seiskari2024gaussian}}
	\label{fig:rolling_shutter}
\end{figure}

To demonstrate the effectiveness of the adaptive distortion-aware transformation in correcting nonlinear distortions, we conducted experiments on a dataset with rolling shutter effects~\cite{seiskari2024gaussian}. As shown in~\cref{fig:rolling_shutter}, while 3DGS~\cite{kerbl20233d} suffers from significant geometric distortions, our method produces results closely matching the ground truth. This capability stems from the inherent flexibility of the adaptive distortion-aware transformation, which can accommodate both linear transformations such as affine transformations and nonlinear ones such as deformable transformations. Thus, as shown in~\cref{tab:ablation}, combining our proposed transformation with rigid body transformations improves performance by addressing the subtle distortions present in motion-blurred images, effectively enhancing the accuracy of the results.

\section{Conclusion}
\label{sec:conclusion}

We propose CRiM-GS, a novel method for reconstructing sharp 3D scenes from blurry images caused by camera motion. Under the assumption that objects remain static during camera movement, we introduce a rigid body transformation method and further enhance it with an adaptive distortion-aware transformation to correct potential real-world distortions. These transformations are continuously modeled using neural ODE in 3D physical space, capturing continuous camera motion trajectories. Drawing inspiration from traditional image deblurring techniques and prior 3D scene deblurring research, we incorporate a pixel-wise weighting strategy with a lightweight CNN. CRiM-GS surpasses state-of-the-art methods in 3D scene deblurring, with extensive experiments validating the effectiveness of each component.
{
    \small
    \bibliographystyle{ieeenat_fullname}
    \bibliography{main}
}
\appendix
\clearpage
\begin{center}
	\textbf{\Large{Appendix}}
	\vspace{5mm}
\end{center}

\section{Implementation Details}
CRiM-GS is trained for 40k iterations based on Mip-Splatting~\cite{yu2024mip}. We set the number of poses $N$ that constitute the continuous camera trajectory to 9. The embedding function of \cref{sec:rigid} is implemented by $\mathtt{nn.Embedding}$ of PyTorch, and the embedded features have the sizes of hidden state of 64. In addition, the sizes of hidden state of the encoders $\mathcal{E}_{r}$, $\mathcal{E}_{c}$, and the decoders $\mathcal{D}_{r}$, $\mathcal{D}_{c}$ are also 64. The neural derivative function $f$ consists of two parallel single linear layers, where one is designated for the rotation matrix and the other for the translation vector. To ensure nonlinearity in the camera motion within the latent space, we apply the $\mathtt{ReLU}$ activation function to each layer. The structure of $g$ is identical to that of $f$. The CNN $\mathcal{F}$ consists of three convolutional layers with 64 channels with kernel size of 5$\times$5 for the first layer and 3$\times$3 for the rest ones. The pixel-wise weights are obtained by applying a pointwise convolutional layer, and the scalar mask $\mathcal{M}$ is obtained by applying another pointwise convolution to the output of $\mathcal{F}$ and averaging it to the batch axis. For first 1k iterations, Gaussian primitives are roughly trained without rigid body transformation and adaptive distortion-aware transformation. After 1k iterations, those transformations start to be trained without the pixel-wise weight and the scalar mask to allow the initial camera motion path to be sufficiently optimized. After 3k iterations, the pixel-wise weight and the scalar mask start training. We set \(\lambda_{c}\), \(\lambda_{o}\), and \(\lambda_{\mathcal{M}}\) to 0.3, \(10^{-4}\), and \(10^{-3}\) respectively for the objective function. All experiments are conducted on a single NVIDIA RTX 3090 GPU.

\section{Additional Ablation Study}
\paragraph{Neural ODE for Camera Motion.}
We conducted a series of ablative experiments on the neural ODE~\cite{chen2018neural} structure of CRiM-GS and the results are shown in \cref{tab:ode}. Note that all experiments, except for those involving splines, incorporate both rigid body transformation and adaptive distortion-aware transformation.

The first approach explores representing the camera motion trajectory using splines in \textit{SE}(3) space instead of neural ODE. This is the primary contribution of BAD-Gaussian~\cite{zhao2024bad}, and thus, the spline experiment results in \cref{tab:ode} correspond to the  performance of BAD-Gaussian. Since BAD-Gaussian assumes very short exposure times, it effectively captures simple camera movements but tends to oversimplify the trajectory when the camera motion becomes nonlinear due to the longer exposure time. In contrast, the our structure assumes more complex camera motion, providing a more comprehensive representation.

 \begin{table}[t]
	\caption{Structural ablation for neural ODE for camera motion. ``-P'' and ``-L'' denote modeling in the physical space and latent space, respectively.}
	\begin{center}
		\resizebox{\columnwidth}{!}{
			\centering
			\setlength{\tabcolsep}{3pt}
			\begin{tabular}{l||c|c|c|c|c|c}
				\toprule 
				
				\multirow{2}{*}{Methods} 			  	& \multicolumn{3}{c|}{Synthetic Scene Dataset}  	   & \multicolumn{3}{c}{Real-World Scene Dataset}  	\\ \cmidrule{2-7}
				&  ~PSNR$\uparrow$~    &~SSIM$\uparrow$~    &~LPIPS$\downarrow$~ &~PSNR$\uparrow$~    &~SSIM$\uparrow$~    &~LPIPS($\downarrow$)~     	\\ \midrule \midrule
				Spline~\cite{zhao2024bad}		& 21.12	& 0.5852	& 0.1007	& 20.82	& 0.5912	& 0.1000 \\
				MLP					& 29.34	& 0.8644	& 0.0784	& 25.94	& 0.7845	& 0.0945 \\
				RNN-P				& 29.11	& 0.8621	& 0.0604	& 25.50	& 0.7719 	& 0.0877 \\
				RNN-L				& \cellcolor{second!35}29.95	& \cellcolor{second!35}0.8881	& \cellcolor{second!35}0.0507	& \cellcolor{second!35}26.78	& 0.8157	& \cellcolor{second!35}0.0713 \\
				Neural ODE-P 	& 29.74 	& 0.8854	& 0.0542	& 26.57	& \cellcolor{second!35}0.8221	& 0.0794 \\ 
				Neural ODE-L	& \cellcolor{best!25}30.44	& \cellcolor{best!25}0.9095	& \cellcolor{best!25}0.0441	& \cellcolor{best!25}27.33	& \cellcolor{best!25}0.8310	& \cellcolor{best!25}0.0634 \\ \bottomrule
			\end{tabular}
		}
	\end{center}
	\label{tab:ode}
\end{table}

The second approach replaces the sequential modeling of camera pathes via neural ODE with a method using MLP. In this method, the embedded features pass through the encoder and three MLP layers with a hidden state size of 64 to obtain the transformation matrices. However, this approach fails to consider the continuous nature of camera motion, resulting in lower performance compared to other methods.

The third and fourth approaches involve using Recurrent Neural Networks (RNN). In the first RNN-based method, the model directly outputs the physical components of the transformation matrices (\textit{e.g.,} components of screw axis). In the second, similar to the proposed method, the RNN extracts features in the latent space, and the physical components are obtained through a decoder. While the latent RNN performs better than the physical space RNN, both are relatively vulnerable to non-uniform blur, as they apply continuous dynamics uniformly over \textit{the same time steps}.

Our approach, which models the camera trajectory using neural ODE, is robust against most types of camera motion blur, as it applies continuous dynamics over irregular time steps. Furthermore, similar to the RNN-based methods, we conduct ablations on the neural ODE in physical and latent spaces, with results indicating higher performance when implemented in the latent space. When neural ODE is applied in the physical space, the features extracted by the encoder (\textit{i.e.,} initial value $\mathbf{z}(t_0)$) directly correspond to physical meanings. However, the components of the screw axis in the rigid body transformation do not exhibit physical continuity, which is why the neural ODE implementation in the latent space achieves the highest performance.

\begin{figure*}[t]
	\centering
	\includegraphics[width=0.9\linewidth]{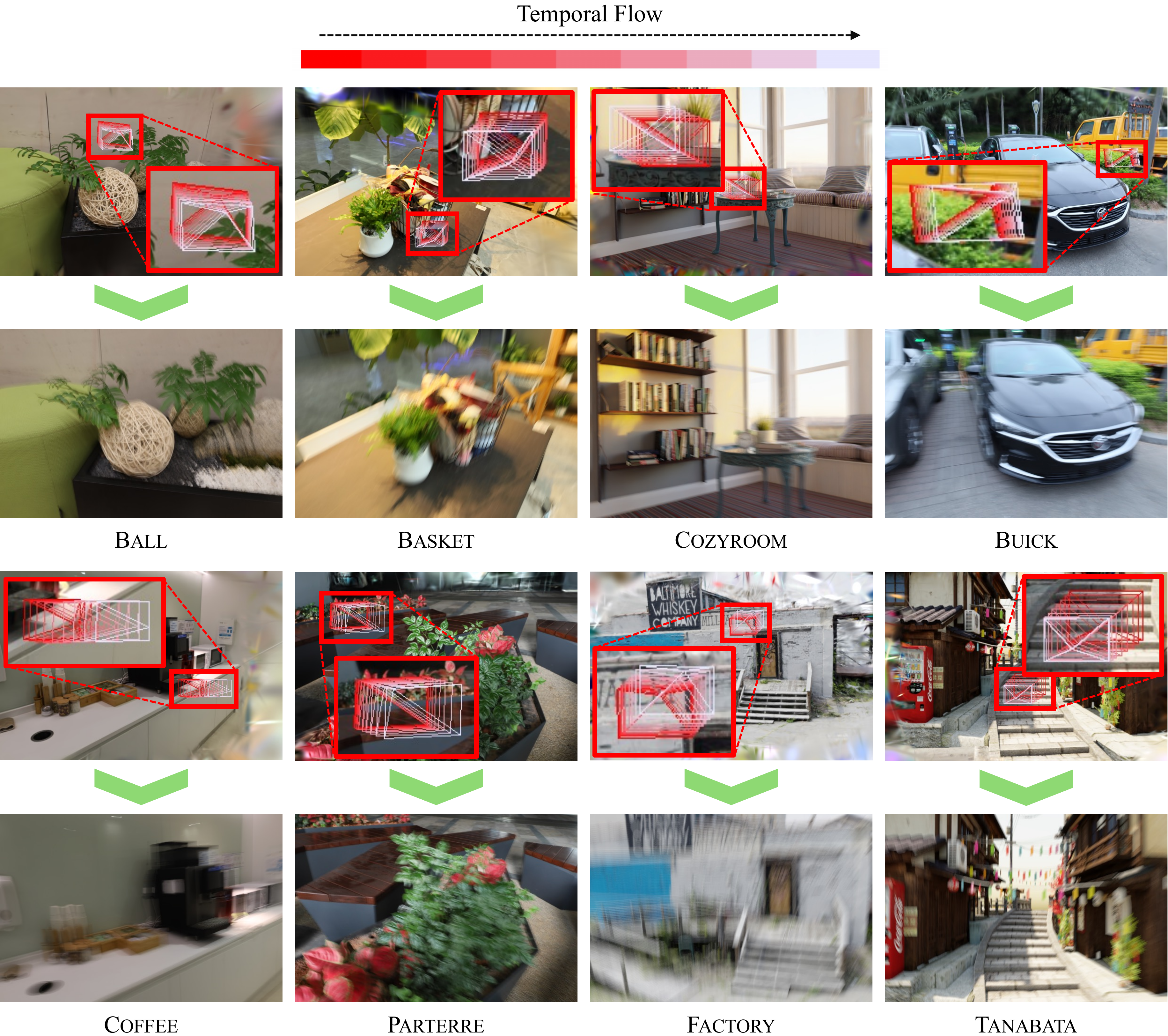}
	\caption{Camera motion trajectory predicted by CRiM-GS for input images with significant blur.}
	\label{fig:kernel_visualize_extreme}
\end{figure*}

\begin{figure*}[t]
	\centering
	\includegraphics[width=0.9\linewidth]{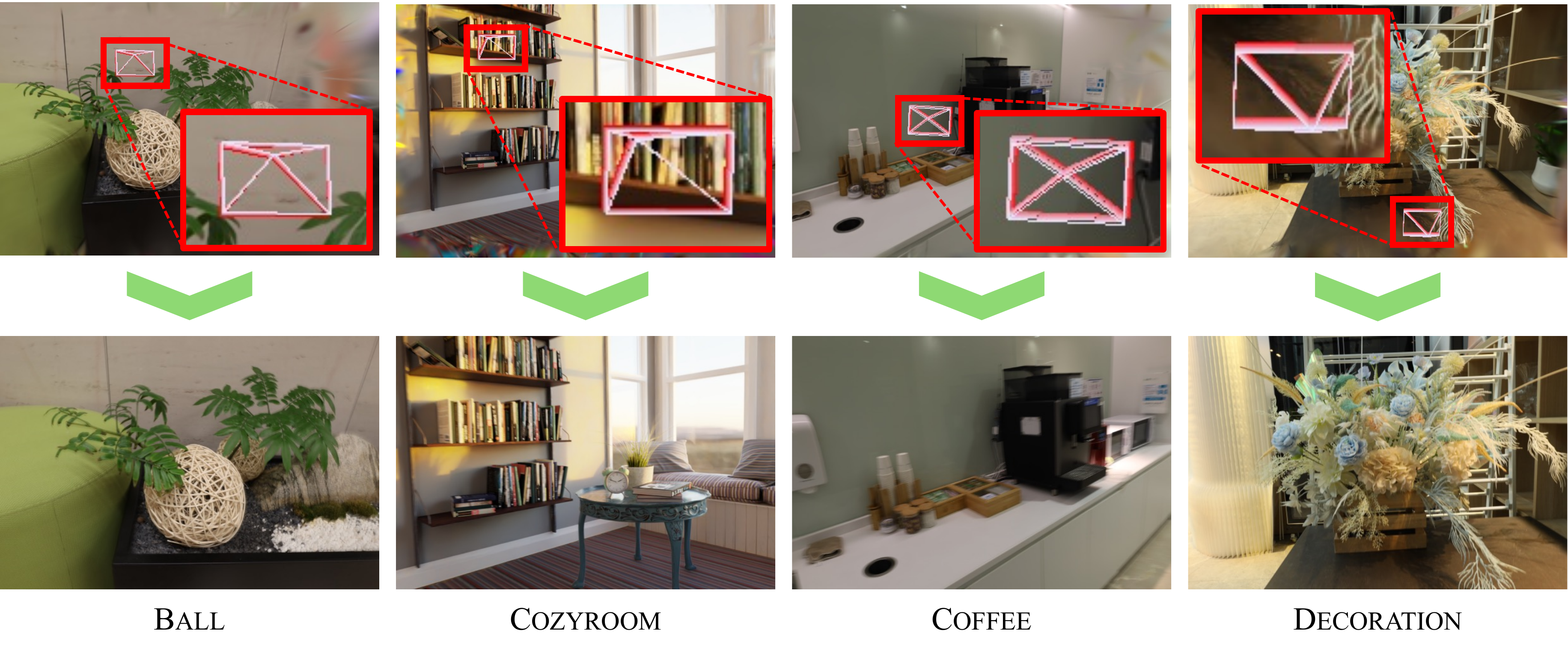}
	\caption{Camera motion trajectory predicted by CRiM-GS for input images with moderate blur.}
	\label{fig:kernel_visualize_moderate}
\end{figure*}

\paragraph{Number of Poses on Camera Motion}
We conduct ablation experiments on the number of camera poses composing the continuous camera motion, and the results are shown in \cref{tab:num_pose}. The results indicate that performance improves across all metrics as $N$ increases, suggesting that a larger number of camera poses allows for more precise modeling of the camera motion trajectory. However, beyond $N=8$, the performance stabilizes and based on these observations, we adopt $N=9$ as it offers the best overall performance.

\begin{table}[t]
	\caption{Experimental results based on the number of poses $N$ along the camera motion trajectory.}
	\vspace{-3mm}
	\begin{center}
		\resizebox{\columnwidth}{!}{
			\centering
			\setlength{\tabcolsep}{2pt}
			\begin{tabular}{l||c|c|c|c|c|c}
				\toprule 
				
				\multirow{2}{*}{Methods} 			  	& \multicolumn{3}{c|}{Synthetic Scene Dataset}  	   & \multicolumn{3}{c}{Real-World Scene Dataset}  	\\ \cmidrule{2-7}
				&  ~PSNR$\uparrow$~    &~SSIM$\uparrow$~    &~LPIPS$\downarrow$~ &~PSNR$\uparrow$~    &~SSIM$\uparrow$~    &~LPIPS($\downarrow$)~     	\\ \midrule \midrule
				$N$=5			& 27.40	& 0.8314	& 0.0936	& 26.42	& 0.8040	& 0.0924 \\
				$N$=6		& 28.78	& 0.8610	& 0.711	& 26.77	& 0.8082	& 0.0822\\
				$N$=7		& 29.52	& 0.8758	& 0.0655	& 27.01 	& 0.8145	& 0.0762\\
				$N$=8 	& 30.17 & 0.9002	& 0.0501	& 26.81	& 0.8101	& 0.0699 \\ 
				$N$=9 	& \cellcolor{second!35}30.44	& \cellcolor{second!35}0.9095	& \cellcolor{best!25}0.0441	& \cellcolor{best!25}27.33	& \cellcolor{best!25}0.8310	& \cellcolor{second!35}0.0634 \\ 
				$N$=10	& \cellcolor{best!25}30.58	& \cellcolor{best!25}0.9124	& \cellcolor{second!35}0.0445	& \cellcolor{second!35}27.12	& \cellcolor{second!35}0.8227	& \cellcolor{best!25}0.0630 \\ \bottomrule
			\end{tabular}
		}
	\end{center}
	\label{tab:num_pose}
	\vspace{-7mm}
\end{table}

\section{Difference from SMURF~\cite{lee2024smurf}}
In this section, we compare our approach with SMURF, a methodology for handling the continuous dynamics of camera motion blur. SMURF utilizes neural ODE to warp a given input ray into continuous rays that simulate camera motion. However, its continuous dynamics are applied only in the 2D pixel space, lacking the inclusion of higher-dimensional camera motion in 3D space. Additionally, as SMURF is implemented on Tensorial Radiance Fields (TensoRF)~\cite{chen2022tensorf}, a ray tracing-based method, it exhibits relatively slower training and rendering speeds.

In contrast, our model uses neural ODE to obtain the 3D camera poses which constitue the camera motion trajectory. Our approach incorporates higher-dimensional information compared to SMURF by operating directly in 3D space rather than the 2D pixel space. Furthermore, as our method is implemented on 3DGS~\cite{kerbl20233d}, a rasterization-based method, it ensures faster training and rendering speeds than SMURF.

\section{Post-Training Pose Optimization}
We reproduce the results of DeblurGS~\cite{oh2024deblurgs} as it performs post-training optimization of test camera poses. This process requires test images and input test camera poses to obtain optimal poses, making it difficult to consider a fair comparison. Nevertheless, as shown in \cref{tab:pose}, CRiM-GS achieves better performance with the same pose optimization process as DeblurGS, and even without the post-training optimization, CRiM-GS outperforms the DeblurGS with optimized poses. This indicates that CRiM-GS is robust to inaccurate camera poses. As illustrated in the error maps in \cref{fig:pose}, CRiM-GS not only performs better than DeblurGS but also optimizes poses better than other models without additional training.

\begin{table}[t]
	\caption{Experimental results based on whether pose optimization is performed after training}
	\begin{center}
		\resizebox{\columnwidth}{!}{
			\centering
			\setlength{\tabcolsep}{2pt}
			\begin{tabular}{l||c|c|c|c|c|c|c}
				\toprule 
				
				\multirow{2}{*}{Methods} 			  & \multirow{2}{*}{~P.O.~} 	& \multicolumn{3}{c|}{Synthetic Scene Dataset}  	   & \multicolumn{3}{c}{Real-World Scene Dataset}  	\\ \cmidrule{3-8}
				&  &~PSNR$\uparrow$~    &~SSIM$\uparrow$~    &~LPIPS$\downarrow$~ &~PSNR$\uparrow$~    &~SSIM$\uparrow$~    &~LPIPS($\downarrow$)~     	\\ \midrule \midrule
				
				DeblurGS*~\cite{oh2024deblurgs}					&				& 20.22		& 0.5454	& 0.1042			& 20.48		& 0.5813		& 0.1186 \\
				\textbf{CRiM-GS}											    &                    & \cellcolor{best!25}30.44    & \cellcolor{best!25}0.9095   & \cellcolor{best!25}0.0441 		 & \cellcolor{best!25}27.33       & \cellcolor{best!25}0.8310      & \cellcolor{best!25}0.0634	\\ \midrule
				DeblurGS*~\cite{oh2024deblurgs}					&\ding{51}				& 30.24		& 0.8922	& 0.0637			& 26.07		& 0.8024		& 0.0958 \\			
				\textbf{CRiM-GS}											          &   \ding{51}           & \cellcolor{best!25}30.65    & \cellcolor{best!25}0.9123   & \cellcolor{best!25}0.0435 		 & \cellcolor{best!25}28.34       & \cellcolor{best!25}0.8478      & \cellcolor{best!25}0.0606 	\\ \bottomrule
			\end{tabular}
		}
	\end{center}
	\label{tab:pose}
\end{table}

\section{Blur Kernel Visualization}
We visualize the camera motion trajectories for input blurry images predicted by CRiM-GS in \cref{fig:kernel_visualize_extreme} and \cref{fig:kernel_visualize_moderate}. \cref{fig:kernel_visualize_extreme} illustrates camera motions for images with significant blur, where the predicted trajectories are continuous over time and align precisely with the input images. \cref{fig:kernel_visualize_moderate} depicts camera motions for images with relatively less blur, where the predicted trajectories show minimal movement, yet still match the input images accurately. These results demonstrate that our blurring kernel effectively models precise continuous camera motion.

\section{Derivation of Rigid Body Motion~\cite{lynch2017modernrobotics}}
In this section, we explain the derivation process for \cref{eq:9} and \cref{eq:10} from the main paper. This derivation aims to expand and simplify the process described in Modern Robotics~\cite{lynch2017modernrobotics} for better clarity and accessibility.

The components of a given screw axis include the unit rotation axis $\hat{\omega}\in\mathbb{R}^{3}$ and the translation component $v\in\mathbb{R}^{3}$. The unit rotation axis consists of the angular velocity $\omega$ and the rotation angle $\theta$:
\begin{equation}
	\hat{\omega}=\frac{\omega}{\theta},\quad where\quad||\hat{\omega}||=1.
\end{equation}
We combine the rotation axis $\hat{\omega}$ and the rotation angle $\theta$ to represent an element of the Lie Algebra, $\mathfrak{so}(3)$, which serves as the linear approximation of the rotation matrix. Before proceeding, $\hat{\omega}$ is converted into a $3\times 3$ skew-symmetric matrix $[\hat{\omega}]$ to compactly express the cross-product operation as a matrix multiplication:
\begin{equation}\label{eq:skew}
	[\hat{\omega}]=\left[
		\begin{array}{ccc}
			0 & -\hat{\omega}_{z} & \hat{\omega}_{y} \\
			\hat{\omega}_{z} & 0 & -\hat{\omega}_{x} \\
			-\hat{\omega}_{y} & \hat{\omega}_{x} & 0 \\
		\end{array}	
	\right] 
	\in \mathfrak{so}(3),~where~ [\hat{\omega}]^3 = I.
\end{equation}
Using the skew-symmetric matrix $[\hat{\omega}]$ and the translation component $v$, the screw axis $[S]$ is expressed. By multiplying this screw axis with $\theta$, we incorporate the magnitude of the rotation and translation along the screw axis:
\begin{equation}
	[\mathcal{S}]\theta=\left[
		\begin{array}{cc}
			[\hat{\omega}]\theta & v\theta \\
			0 & 0 \\
		\end{array}
	\right]
	\in \mathfrak{se}(3),
\end{equation}
where$\mathfrak{se}(3)$ represents the Lie Algebra, which corresponds to the infinitesimal changes of the Lie Group \textit{SE}(3). To map this infinitesimal change to the \textit{SE}(3) transformation matrix $\mathbf{T}=e^{[\mathcal{S}]\theta}$, we use the Taylor expansion, following these steps:\\
{\small{\begin{align}
	e^{[\mathcal{S}]\theta} & = \sum_{n=0}^{\infty} [\mathcal{S}]^{n}\frac{\theta^{n}}{n!} \\
											  & = I + [\mathcal{S}]\theta + [\mathcal{S}]^2\frac{\theta^2}{2!} + \cdots \\
											  	& = \left[\begin{array}{cc}
											  		I + [\hat{\omega}]\theta + [\hat{\omega}]^{2}\frac{\theta^{2}}{2!}+\cdots & \left(I\theta + [\hat{\omega}]\frac{\theta^{2}}{2!}+\cdots\right)v \\
											  		0 & 1 \\
											  	\end{array}
											  	\right] \\
											  	& = \left[\begin{array}{cc}
											  		e^{[\hat{\omega}]\theta} & G(\theta)v \\
											  		0 & 1 \\
											  		\end{array}
											  	\right] \in \textit{SE}(3)\\
											  	& \because~[\mathcal{S}]^{n}=\left[
											  	\begin{array}{cc}
											  		[\hat{\omega}]^{n} & [\hat{\omega}]^{n-1}v \\
											  		0 & 0 \\
											  	\end{array}
											  	\right]
\end{align}}}
For the rotation matrix $e^{[\hat{\omega}]}\theta$, we simplify it using the Taylor expansion and \cref{eq:skew}, resulting in:
{\small{\begin{align}
	e^{[\hat{\omega}]\theta} & = I + [\hat{\omega}]\theta + [\hat{\omega}]^{2}\frac{\theta^{2}}{2!} + [\hat{\omega}]^{3}\frac{\theta^{3}}{3!} +  + [\hat{\omega}]^{4}\frac{\theta^{4}}{4!}\cdots \\
											  & = I + \left(\theta - \frac{\theta^{3}}{3!} + \cdots\right)[\hat{\omega}] + \left(\frac{\theta^{2}}{2!} - \frac{\theta^{4}}{4!} + \cdots\right)[\hat{\omega}]^{2} \\
											  & = I + \sin \theta[\hat{\omega}] + (1 - \cos \theta)[\hat{\omega}]^2 \in \textit{SO}(3)
\end{align}}}
The translational component $G(\theta)$ is also derived using the Taylor expansion and \cref{eq:skew}:
{\small{\begin{align}
	G(\theta) & = I\theta + [\hat{\omega}]\frac{\theta^{2}}{2!} + [\hat{\omega}]^{2}\frac{\theta^{3}}{3!} + [\hat{\omega}]^{3}\frac{\theta^{4}}{4!}+\cdots \\
	& = I\theta + \left(\frac{\theta^{2}}{2!} - \frac{\theta^{4}}{4!} + \cdots\right)[\hat{\omega}] + \left(\frac{\theta^{3}}{3!} - \frac{\theta^{5}}{5!} + \cdots\right)[\hat{\omega}]^{2} \\
	& = I\theta + (1 - \cos \theta)[\hat{\omega}] + (\theta - \sin \theta)[\hat{\omega}]^2
\end{align}}}
The term $G(\theta)$ physically represents the total translational motion caused by the rotational motion as the rigid body rotates by $\theta$. In other words, $G(\theta)$ indicates how rotational motion contributes to translational motion, which can also be expressed as an integral of the rotation motion:
\begin{align}
	G(\theta) & = \int_{0}^{\theta} e^{[\hat{\omega}]\theta}d\theta \\
	& = \int_{0}^{\theta} \left(I + \sin \theta[\hat{\omega}] + (1 - \cos \theta)[\hat{\omega}]^2\right)d\theta \\
	& = I\theta + (1 - \cos \theta)[\hat{\omega}] + (\theta - \sin \theta)[\hat{\omega}]^2
\end{align}
Through the above process, we derive \cref{eq:9} and \cref{eq:10} in the main paper, improving readability of the paper and providing a clear foundation for understanding the mathematical framework.

\section{Per-Scene Quantitative Results}
We show the per-scene quantitative performance on synthetic and real-world datasets in \cref{tab:synthetic} and \cref{tab:real}. CRiM-GS demonstrates superior performance on most scenes in the real-world dataset and achieves the highest performance across all scenes in the synthetic dataset. Notably, CRiM-GS achieves the best LPIPS scores for all scenes, highlighting the superior quality achieved by our CRiM-GS.

\section{Additional Qualitative Results}
\begin{figure*}[t]
	\centering
	\includegraphics[width=\linewidth]{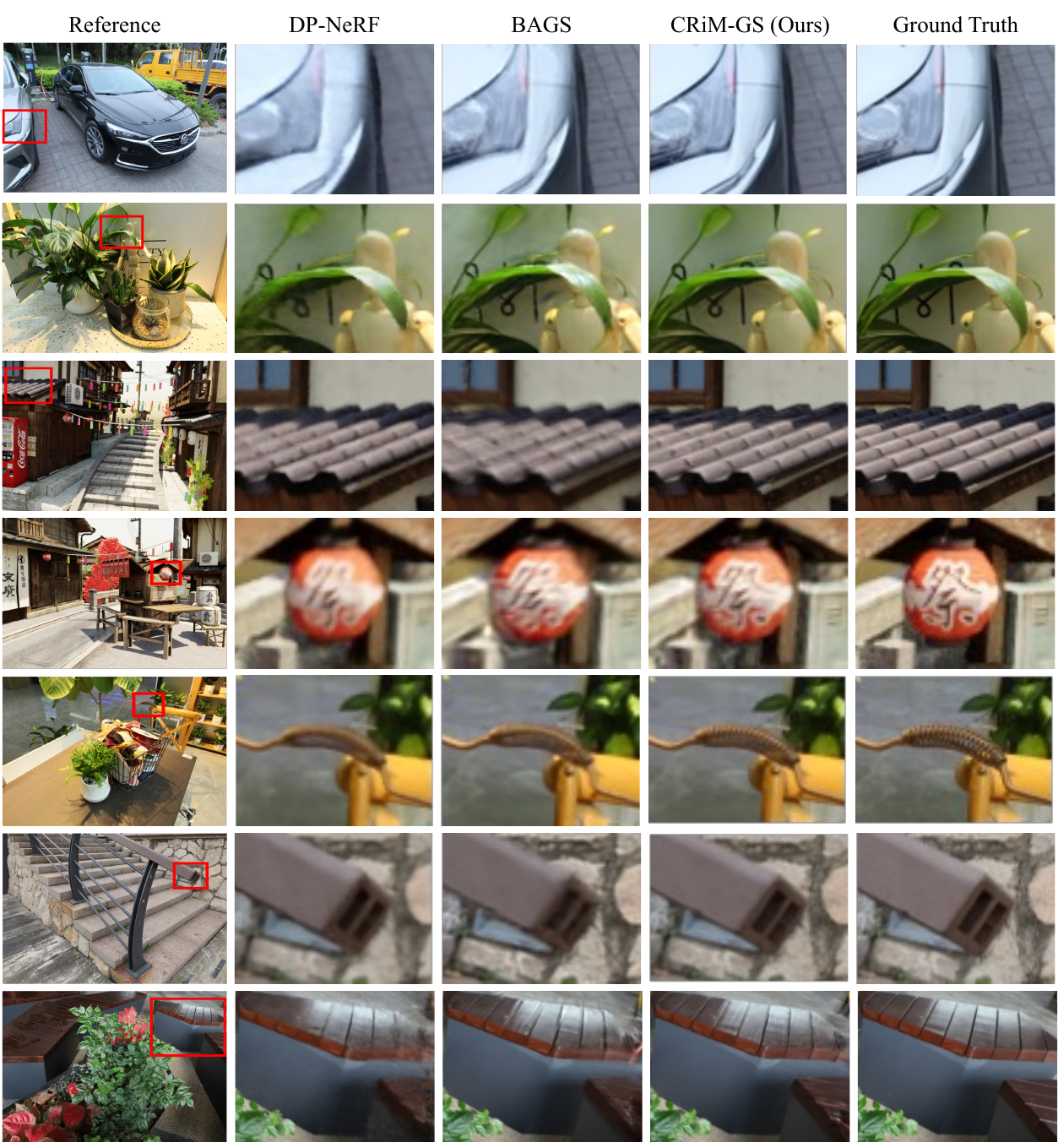}
	\caption{Additional Qualitative Comparison on the Synthetic and Real-World Scenes.}
	\label{fig:additional}
\end{figure*}

We provide additional visualization results in \cref{fig:additional}, which demonstrate that our CRiM-GS outperforms not only in quantitative metrics but also in qualitative performance. For comparative videos, please refer to \textit{the supplementary materials}.

\begin{figure*}[t]
	\centering
	\includegraphics[width=\linewidth]{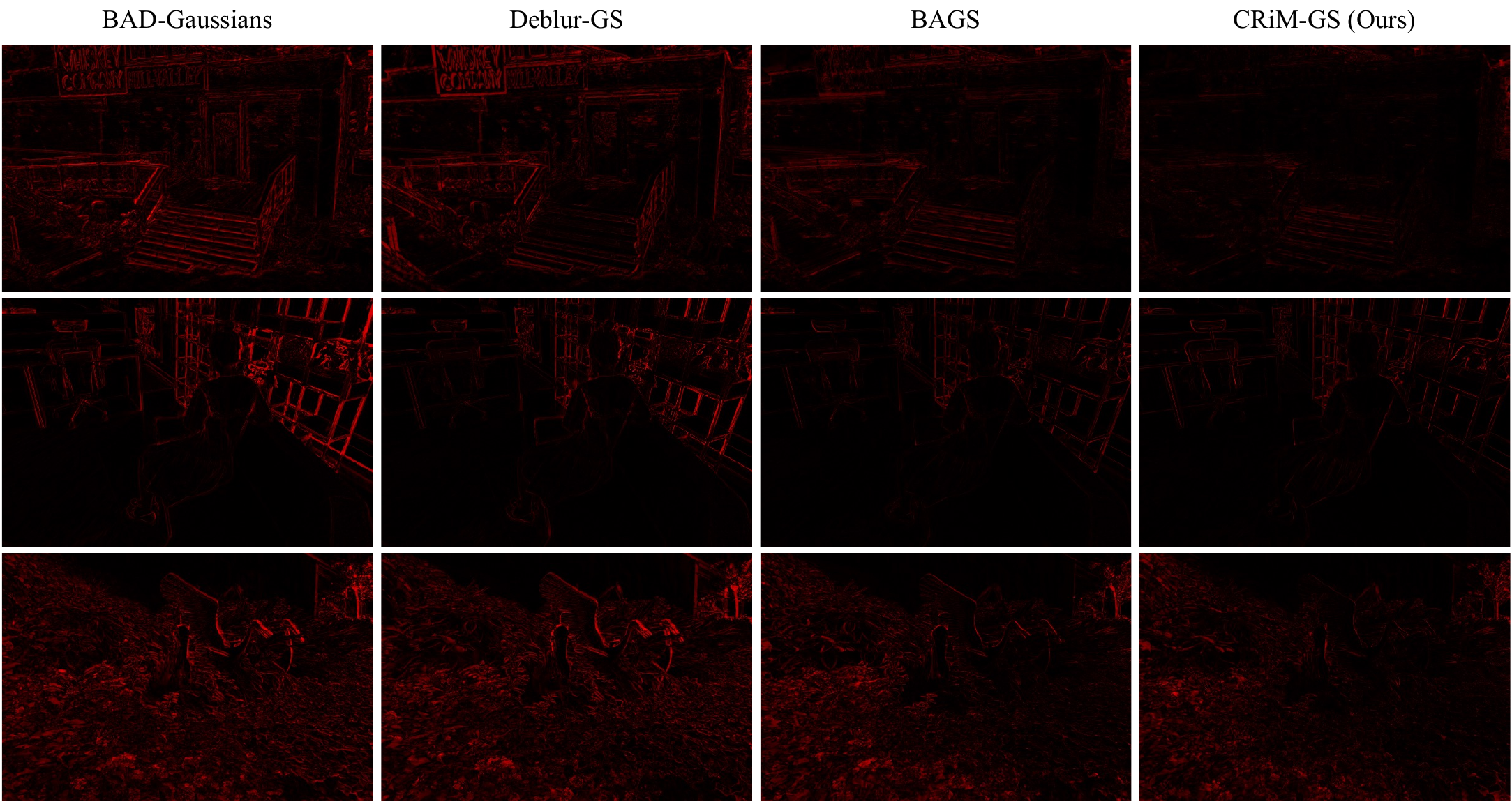}
	\caption{Error Map Comparison.}
	\label{fig:pose}
\end{figure*}

\paragraph{Error Map Visualization.}
We extract error maps for three scenes to enable direct comparison with other methods, and the results are shown in \cref{fig:pose}. While BAD-Gaussians~\cite{zhao2024bad} and DeblurGS~\cite{oh2024deblurgs} exhibit relatively high errors, BAGS~\cite{peng2024bags} and CRiM-GS demonstrate lower errors. Notably, our model captures the overall contours of the scene more effectively than BAGS.

\section{Limitation}
Despite achieving superior qualitative and quantitative performance with fast rendering, the proposed model faces room for optimization in terms of training efficiency. The primary challenge lies in the multiple rendering steps per iteration and the use of the CNN module $\mathcal{F}$ for pixel-wise weighted sums, which is employed only during training. However, this aspect offers opportunities for future improvements. Designing continuous motion directly at the 3D Gaussian level could streamline the process, reducing training time without compromising performance.

\begin{table*}[t]
	\caption{Per-Scene Quantitative Performance on the Real-World Scenes.}
	\begin{center}
		\resizebox{\linewidth}{!}{
			\centering
			\setlength{\tabcolsep}{1.5pt}
			\scriptsize
			\begin{tabular}{l||c|c|c|c|c|c|c|c|c|c|c|c|c|c|c}
				\toprule 
				
				\multirow{2}{*}{Real-World Scene\ } 			   & \multicolumn{3}{c|}{\textsc{Ball}}  	   & \multicolumn{3}{c|}{\textsc{Basket}}  	   & \multicolumn{3}{c|}{\textsc{Buick}}  				& \multicolumn{3}{c|}{\textsc{Coffee}} 		 & \multicolumn{3}{c}{\textsc{Decoration}}  \\ \cmidrule{2-16}
				&\ PSNR$\uparrow$ \     &\ SSIM$\uparrow$ \     &\ LPIPS$\downarrow$ \  &\ PSNR$\uparrow$ \     &\ SSIM$\uparrow$ \     &\ LPIPS$\downarrow$ \ 		&\ PSNR$\uparrow$ \     &\ SSIM$\uparrow$ \     &\ LPIPS$\downarrow$ \ 		&\ PSNR$\uparrow$ \     &\ SSIM$\uparrow$ \       &\ LPIPS$\downarrow$ \ 	&\ PSNR$\uparrow$ \     &\ SSIM$\uparrow$ \       &\ LPIPS$\downarrow$ \   	\\ \midrule \midrule
				Naive NeRF~\cite{mildenhall2020nerf}                            				 & 24.08    & 0.6237   & 0.3992 		& 23.72       & 0.7086      & 0.3223      & 21.59    & 0.6325      & 0.3502    		& 26.48    & 0.8064      & 0.2896    		& 22.39    & 0.6609     & 0.3633    \\ \midrule
				Deblur-NeRF~\cite{ma2022deblurnerf}                         				& 27.36    & 0.7656   & 0.2230 		 & 27.67       & 0.8449      & 0.1481        & 24.77     & 0.7700      & 0.1752    		& 30.93     & 0.8981      & 0.1244    		 & 24.19    & 0.7707     & 0.1862    \\
				PDRF-10*~\cite{peng2023pdrf}                            				 & 27.37    & 0.7642   & 0.2093 		& 28.36       & 0.8736      & 0.1179      & 25.73    & 0.7916      & 0.1582    		& \cellcolor{second!35}31.79    & 0.9002      & 0.1133    		& 23.55   & 0.7508     & 0.2145    \\ 
				BAD-NeRF*~\cite{wang2023bad}                            				 & 21.33 & 0.5096 & 0.4692 			& 26.44 		& 0.8080 	& 0.1325 		& 21.63  & 0.6429 		& 0.2593 		& 28.98 & 0.8369 		& 0.1956 		& 22.13	 	& 0.6316 	& 0.2894    \\ 
				DP-NeRF~\cite{lee2023dp}                        						  & 27.20    & 0.7652   & 0.2088 		& 27.74   	 & 0.8455      & 0.1294    		& 25.70    & 0.7922      & 0.1405    	& 31.19    & 0.9049      & 0.1002    		& 24.31    & 0.7811     & 0.1639   \\ \midrule
				DeblurGS~\cite{oh2024deblurgs}											& 21.44		& 0.5760		& 0.1480		& 21.00		& 0.6706		& 0.0908			& 18.91		& 0.5539		& 0.1442	& 25.97	& 0.8299		& \cellcolor{second!35}0.0614		& 19.60	& 0.5713		& 0.1035 \\
				BAD-Gaussians*~\cite{zhao2024bad}                            				 & 22.14    & 0.5791   & \cellcolor{second!35}0.1095 		& 21.48       & 0.6787      & \cellcolor{second!35}0.0603      & 18.81    & 0.5271      & 0.0952    		& 24.54    & 0.7627      & 0.1031    		& 20.30   & 0.6162     & 0.0881    \\ 
				Deblurring 3DGS~\cite{lee2024deblurring}												& 28.27 & 0.8233 & 0.1413 & 28.42 & 0.8713 & 0.1155		& 25.95		& 0.8367		& 0.0954		& 32.84		& 0.9312 & 	0.0676	& 25.87 & 0.8540 &  0.0933 \\
				BAGS~\cite{peng2024bags}                            				 & \cellcolor{second!35}27.68    & \cellcolor{second!35}0.7990   & 0.1500 		& \cellcolor{second!35}29.54       & \cellcolor{second!35}0.9000      & 0.0680      & \cellcolor{second!35}26.18    & \cellcolor{best!25}0.8440      & \cellcolor{second!35}0.0880    		& 31.59    & \cellcolor{second!35}0.9080      & 0.0960    		& \cellcolor{second!35}26.09   & \cellcolor{second!35}0.8580     & \cellcolor{second!35}0.0830    \\ \midrule
				\textbf{CRiM-GS}                            				 & \cellcolor{best!25}28.25    & \cellcolor{best!25}0.8064   & \cellcolor{best!25}0.1074 		& \cellcolor{best!25}30.85       & \cellcolor{best!25}0.9048      & \cellcolor{best!25}0.0425      & \cellcolor{best!25}26.51    & \cellcolor{second!35}0.8379      & \cellcolor{best!25}0.0585    		& \cellcolor{best!25}32.26    & \cellcolor{best!25}0.9230      & \cellcolor{best!25}0.0421    		& \cellcolor{best!25}26.34   & \cellcolor{best!25}0.8610     & \cellcolor{best!25}0.0569    \\ \midrule \midrule
				
				\multirow{2}{*}{Real-World Scene\ } 			   & \multicolumn{3}{c|}{\textsc{Girl}}  	   & \multicolumn{3}{c|}{\textsc{Heron}}  	   & \multicolumn{3}{c|}{\textsc{Parterre}}  				& \multicolumn{3}{c|}{\textsc{Puppet}} 		 & \multicolumn{3}{c}{\textsc{Stair}}  \\ \cmidrule{2-16}
				&\ PSNR$\uparrow$ \     &\ SSIM$\uparrow$ \     &\ LPIPS$\downarrow$ \  &\ PSNR$\uparrow$ \     &\ SSIM$\uparrow$ \     &\ LPIPS$\downarrow$ \ 		&\ PSNR$\uparrow$ \     &\ SSIM$\uparrow$ \     &\ LPIPS$\downarrow$ \ 		&\ PSNR$\uparrow$ \     &\ SSIM$\uparrow$ \       &\ LPIPS$\downarrow$ \ 	&\ PSNR$\uparrow$ \     &\ SSIM$\uparrow$ \       &\ LPIPS$\downarrow$ \   	\\ \midrule \midrule
				Naive NeRF~\cite{mildenhall2020nerf}                            				 & 20.07    & 0.7075   & 0.3196 		& 20.50       & 0.5217      & 0.4129      & 23.14    & 0.6201      & 0.4046    		& 22.09    & 0.6093      & 0.3389    		& 22.87    & 0.4561     & 0.4868    \\ \midrule
				Deblur-NeRF~\cite{ma2022deblurnerf}                         				& 22.27    & 0.7976   & 0.1687 		 & 22.63       & 0.6874      & 0.2099        & 25.82     & 0.7597      & 0.2161    		& 25.24     & 0.7510      & 0.1577    		 & 25.39    & 0.6296     & 0.2102    \\
				PDRF-10*~\cite{peng2023pdrf}                            				 & 24.12    & 0.8328   & 0.1679 		& 22.53       & 0.6880      & 0.2358      & 25.36    & 0.7601      & 0.2263    		& 25.02    & 0.7496     & 0.1532    		& 25.20    & 0.6235     & 0.2288    \\ 
				BAD-NeRF*~\cite{wang2023bad}                            				 & 18.10 & 0.5652 	& 0.3933 		& 22.18 		& 0.6479 	& 0.2226 		& 23.44 	& 0.6243 	& 0.3151 		& 22.48 	& 0.6249 	& 0.2762 			& 21.52 	& 0.4237 	& 0.3341    \\ 
				DP-NeRF~\cite{lee2023dp}                        						  & 23.33    & 0.8139   & 0.1498 		& \cellcolor{best!25}22.88   	 & 0.6930      & 0.1914    		& 25.86    & 0.7665      & 0.1900    	& 25.25    & 0.7536      & 0.1505    		& 25.59    & 0.6349      & 0.1772   \\ \midrule 
				DeblurGS*~\cite{oh2024deblurgs}												& 18.28		& 0.6660	& 0.1044		& 18.70	& 0.4579	& 0.1488		& 19.78		& 0.4953	& 0.1744	& 19.50	& 0.5302	& 0.1297		& 21.63	& 0.4623		& 0.0808	\\
				BAD-Gaussians*~\cite{zhao2024bad}                            				 	& 18.72    & 0.6765   & 0.0925 		& 18.53       & 0.4498      & 0.1412      & 20.79    & 0.5589      & 0.1298   		& 21.08    & 0.6036     & 0.1022    		& 21.77    & 0.4640     & \cellcolor{second!35}0.0782    \\ 
				Deblurring 3DGS~\cite{lee2024deblurring}							& 23.26 & 0.8390 & 	0.1011	& 23.14 & 0.7438 & 0.1543	 		& 26.17 & 0.8144	& 0.1206		& 25.67 & 0.8051	& 0.0941		& 26.46 & 0.7050 & 0.1123 \\
				BAGS~\cite{peng2024bags}                            				 		& \cellcolor{second!35}25.45    & \cellcolor{second!35}0.8690   & \cellcolor{second!35}0.0790 		& 22.04       & \cellcolor{second!35}0.7150      & \cellcolor{second!35}0.1260      & \cellcolor{second!35}25.92    & \cellcolor{best!25}0.8190      & \cellcolor{second!35}0.0920    		& \cellcolor{second!35}25.81    & \cellcolor{second!35}0.8040     & \cellcolor{second!35}0.0940    		& \cellcolor{second!35}26.69    & \cellcolor{second!35}0.7210     & 0.0800    \\ \midrule
				\textbf{CRiM-GS}                            				 	& \cellcolor{best!25}26.20    & \cellcolor{best!25}0.8780   & \cellcolor{best!25}0.0438 		& \cellcolor{second!35}22.84       & \cellcolor{best!25}0.7234      & \cellcolor{best!25}0.1110     & \cellcolor{best!25}26.09    & \cellcolor{second!35}0.8126      & \cellcolor{best!25}0.0698    		& \cellcolor{best!25}26.66    & \cellcolor{best!25}0.8236     & \cellcolor{best!25}0.0613    		& \cellcolor{best!25}27.30    & \cellcolor{best!25}0.7390     & \cellcolor{best!25}0.0415    \\ \bottomrule
			\end{tabular}
		}
	\end{center}
	\label{tab:real}
\end{table*}

\begin{table*}[t]
	\caption{Per-Scene Quantitative Performance on the Synthetic Scenes.}
	\begin{center}
		\resizebox{\linewidth}{!}{
			\centering
			\setlength{\tabcolsep}{1.5pt}
			\scriptsize
			\begin{tabular}{l||c|c|c|c|c|c|c|c|c|c|c|c|c|c|c}
				\toprule 
				
				\multirow{2}{*}{Synthetic Scene} 			   & \multicolumn{3}{c|}{\textsc{Factory}}  	   & \multicolumn{3}{c|}{\textsc{CozyRoom}}  	   & \multicolumn{3}{c|}{\textsc{Pool}}  				& \multicolumn{3}{c|}{\textsc{Tanabata}} 		 & \multicolumn{3}{c}{\textsc{Trolley}}  \\ \cmidrule{2-16}
				&\ PSNR$\uparrow$ \     &\ SSIM$\uparrow$ \     &\ LPIPS$\downarrow$ \  &\ PSNR$\uparrow$ \     &\ SSIM$\uparrow$ \     &\ LPIPS$\downarrow$ \ 		&\ PSNR$\uparrow$ \     &\ SSIM$\uparrow$ \     &\ LPIPS$\downarrow$ \ 		&\ PSNR$\uparrow$ \     &\ SSIM$\uparrow$ \       &\ LPIPS$\downarrow$ \ 	&\ PSNR$\uparrow$ \     &\ SSIM$\uparrow$ \       &\ LPIPS$\downarrow$ \   	\\ \midrule \midrule
				Deblur-NeRF~\cite{ma2022deblurnerf}                         				& 25.60    & 0.7750   & 0.2687 		 & 32.08       & 0.9261      & 0.0477        & 31.61     & 0.8682      & 0.1246    		& 27.11     & 0.8640      & 0.1228    		 & 27.45    & 0.8632     & 0.1363    \\
				PDRF-10*~\cite{peng2023pdrf}                         					   & 25.87    & \cellcolor{second!35}0.8316   & 0.1915 		 		  & 31.13       &  0.9225     & 0.0439        			& 31.00     & 0.8583      & 0.1408    				& \cellcolor{second!35}28.01     & \cellcolor{second!35}0.8931      & 0.1004    		 		 & \cellcolor{second!35}28.29    & \cellcolor{second!35}0.8921     & 0.0931 		    \\
				BAD-NeRF*~\cite{wang2023bad}                         				& 24.43    & 0.7274   & 0.2134 		 & 29.77       & 0.8864      & 0.0616        & 31.51     & 0.8620      & 0.0802    		& 25.32     & 0.8081      & 0.1077    		 & 25.58    & 0.8049     & 0.1008    \\
				DP-NeRF~\cite{lee2023dp}                        				 	 & \cellcolor{second!35}25.91    & 0.7787   & 0.2494 		& \cellcolor{second!35}32.65   	 & 0.9317      & 0.0355    		& \cellcolor{second!35}31.96    & \cellcolor{second!35}0.8768      & 0.0908    	& 27.61    & 0.8748      & 0.1033    		& 28.03    & 0.8752      & 0.1129   \\ \midrule
				DeblurGS*~\cite{oh2024deblurgs}											& 17.82		& 0.3980	& \cellcolor{second!35}0.1453		& 22.88		& 0.6758	& 0.0514		& 24.53		& 0.6095	& 0.1171		& 17.54		& 0.4800		& 0.1132		& 18.32		& 0.5638		& 0.0942 \\
				BAD-Gaussians*~\cite{zhao2024bad}										 	 & 16.84    & 0.3220   & 0.2425 			& 22.31       & 0.6954      & 0.0519      & 26.70    & 0.7051      & \cellcolor{second!35}0.0642    		& 20.13    & 0.6095      & \cellcolor{second!35}0.0729    		& 19.60    & 0.5941     & \cellcolor{second!35}0.0722    \\ 
				Deblurring 3DGS~\cite{lee2024deblurring}												& 24.01    & 0.7333   & 0.2326 			& 31.45       & 0.9222      & 0.0367      & 31.87    & 0.8829      & 0.0751    		& 27.01    & 0.8807      & 0.0785    		& 26.88    & 0.8710     & 0.1028    \\
				BAGS~\cite{peng2024bags}												& 22.35    & 0.6639   & 0.2277 			& 32.21       & \cellcolor{second!35}0.9359      & \cellcolor{second!35}0.0245      & 28.72    & 0.8404      & 0.0804    		& 26.79    & 0.8735      & 0.1099    		& 26.61    & 0.8627     & 0.1156    \\ \midrule \midrule
				\textbf{CRiM-GS}											 & \cellcolor{best!25}28.72    & \cellcolor{best!25}0.8874   & \cellcolor{best!25}0.0604 			& \cellcolor{best!25}33.01       & \cellcolor{best!25}0.9409      & \cellcolor{best!25}0.0209      & \cellcolor{best!25}32.05    & \cellcolor{best!25}0.8882      & \cellcolor{best!25}0.0524    		& \cellcolor{best!25}29.04    & \cellcolor{best!25}0.9180      & \cellcolor{best!25}0.0376    		& \cellcolor{best!25}29.36    & \cellcolor{best!25}0.9131     & \cellcolor{best!25}0.0492    \\  \bottomrule
			\end{tabular}
		}
		\label{tab:synthetic}
	\end{center}
	\vspace{-5mm}
\end{table*}

\end{document}


\clearpage
\setcounter{page}{1}
\maketitlesupplementary

\section{Implementation Details}
CRiM-GS is trained for 40k iterations based on Mip-Splatting~\cite{yu2024mip}. We set the number of poses $N$ that constitute the continuous camera trajectory to 9. The embedding function of Sec.~\textcolor{red}{4.2} of the main paper is implemented by $\mathtt{nn.Embedding}$ of PyTorch, and the embedded features have the sizes of hidden state of 64. In addition, the sizes of hidden state of the encoders $\mathcal{E}_{r}$, $\mathcal{E}_{c}$, and the decoders $\mathcal{D}_{r}$, $\mathcal{D}_{c}$ are also 64. The neural derivative function $f$ consists of two parallel single linear layers, where one is designated for the rotation matrix and the other for the translation vector. To ensure nonlinearity in the camera motion within the latent space, we apply the $\mathtt{ReLU}$ activation function to each layer. The structure of $g$ is identical to that of $f$. The CNN $\mathcal{F}$ consists of three convolutional layers with 64 channels with kernel size of 5$\times$5 for the first layer and 3$\times$3 for the rest ones. The pixel-wise weights are obtained by applying a pointwise convolutional layer, and the scalar mask $\mathcal{M}$ is obtained by applying another pointwise convolution to the output of $\mathcal{F}$ and averaging it to the batch axis. For first 1k iterations, Gaussian primitives are roughly trained without rigid body transformation and adaptive distortion-aware transformation. After 1k iterations, those transformations start to be trained without the pixel-wise weight and the scalar mask to allow the initial camera motion path to be sufficiently optimized. After 3k iterations, the pixel-wise weight and the scalar mask start training. We set \(\lambda_{c}\), \(\lambda_{o}\), and \(\lambda_{\mathcal{M}}\) to 0.3, \(10^{-4}\), and \(10^{-3}\) respectively for the objective function. All experiments are conducted on a single NVIDIA RTX 3090 GPU.

\section{Additional Ablation Study}
\paragraph{Neural ODE for Camera Motion.}
We conducted a series of ablative experiments on the neural ODE~\cite{chen2018neural} structure of CRiM-GS and the results are shown in \cref{tab:ode}. Note that all experiments, except for those involving splines, incorporate both rigid body transformation and adaptive distortion-aware transformation.

The first approach explores representing the camera motion trajectory using splines in \textit{SE}(3) space instead of neural ODE. This is the primary contribution of BAD-Gaussian~\cite{zhao2024bad}, and thus, the spline experiment results in \cref{tab:ode} correspond to the  performance of BAD-Gaussian. Since BAD-Gaussian assumes very short exposure times, it effectively captures simple camera movements but tends to oversimplify the trajectory when the camera motion becomes nonlinear due to the longer exposure time. In contrast, the our structure assumes more complex camera motion, providing a more comprehensive representation.

 \begin{table}[t]
	\caption{Structural ablation for neural ODE for camera motion. ``-P'' and ``-L'' denote modeling in the physical space and latent space, respectively.}
	\begin{center}
		\resizebox{\columnwidth}{!}{
			\centering
			\setlength{\tabcolsep}{3pt}
			\begin{tabular}{l||c|c|c|c|c|c}
				\toprule 
				
				\multirow{2}{*}{Methods} 			  	& \multicolumn{3}{c|}{Synthetic Scene Dataset}  	   & \multicolumn{3}{c}{Real-World Scene Dataset}  	\\ \cmidrule{2-7}
				&  ~PSNR$\uparrow$~    &~SSIM$\uparrow$~    &~LPIPS$\downarrow$~ &~PSNR$\uparrow$~    &~SSIM$\uparrow$~    &~LPIPS($\downarrow$)~     	\\ \midrule \midrule
				Spline~\cite{zhao2024bad}		& 21.12	& 0.5852	& 0.1007	& 20.82	& 0.5912	& 0.1000 \\
				MLP					& 29.34	& 0.8644	& 0.0784	& 25.94	& 0.7845	& 0.0945 \\
				RNN-P				& 29.11	& 0.8621	& 0.0604	& 25.50	& 0.7719 	& 0.0877 \\
				RNN-L				& \cellcolor{second!35}29.95	& \cellcolor{second!35}0.8881	& \cellcolor{second!35}0.0507	& \cellcolor{second!35}26.78	& 0.8157	& \cellcolor{second!35}0.0713 \\
				Neural ODE-P 	& 29.74 	& 0.8854	& 0.0542	& 26.57	& \cellcolor{second!35}0.8221	& 0.0794 \\ 
				Neural ODE-L	& \cellcolor{best!25}30.44	& \cellcolor{best!25}0.9095	& \cellcolor{best!25}0.0441	& \cellcolor{best!25}27.33	& \cellcolor{best!25}0.8310	& \cellcolor{best!25}0.0634 \\ \bottomrule
			\end{tabular}
		}
	\end{center}
	\label{tab:ode}
\end{table}

The second approach replaces the sequential modeling of camera pathes via neural ODE with a method using MLP. In this method, the embedded features pass through the encoder and three MLP layers with a hidden state size of 64 to obtain the transformation matrices. However, this approach fails to consider the continuous nature of camera motion, resulting in lower performance compared to other methods.

The third and fourth approaches involve using Recurrent Neural Networks (RNN). In the first RNN-based method, the model directly outputs the physical components of the transformation matrices (\textit{e.g.,} components of screw axis). In the second, similar to the proposed method, the RNN extracts features in the latent space, and the physical components are obtained through a decoder. While the latent RNN performs better than the physical space RNN, both are relatively vulnerable to non-uniform blur, as they apply continuous dynamics uniformly over \textit{the same time steps}.

Our approach, which models the camera trajectory using neural ODE, is robust against most types of camera motion blur, as it applies continuous dynamics over irregular time steps. Furthermore, similar to the RNN-based methods, we conduct ablations on the neural ODE in physical and latent spaces, with results indicating higher performance when implemented in the latent space. When neural ODE is applied in the physical space, the features extracted by the encoder (\textit{i.e.,} initial value $\mathbf{z}(t_0)$) directly correspond to physical meanings. However, the components of the screw axis in the rigid body transformation do not exhibit physical continuity, which is why the neural ODE implementation in the latent space achieves the highest performance.

\begin{figure*}[t]
	\centering
	\includegraphics[width=0.9\linewidth]{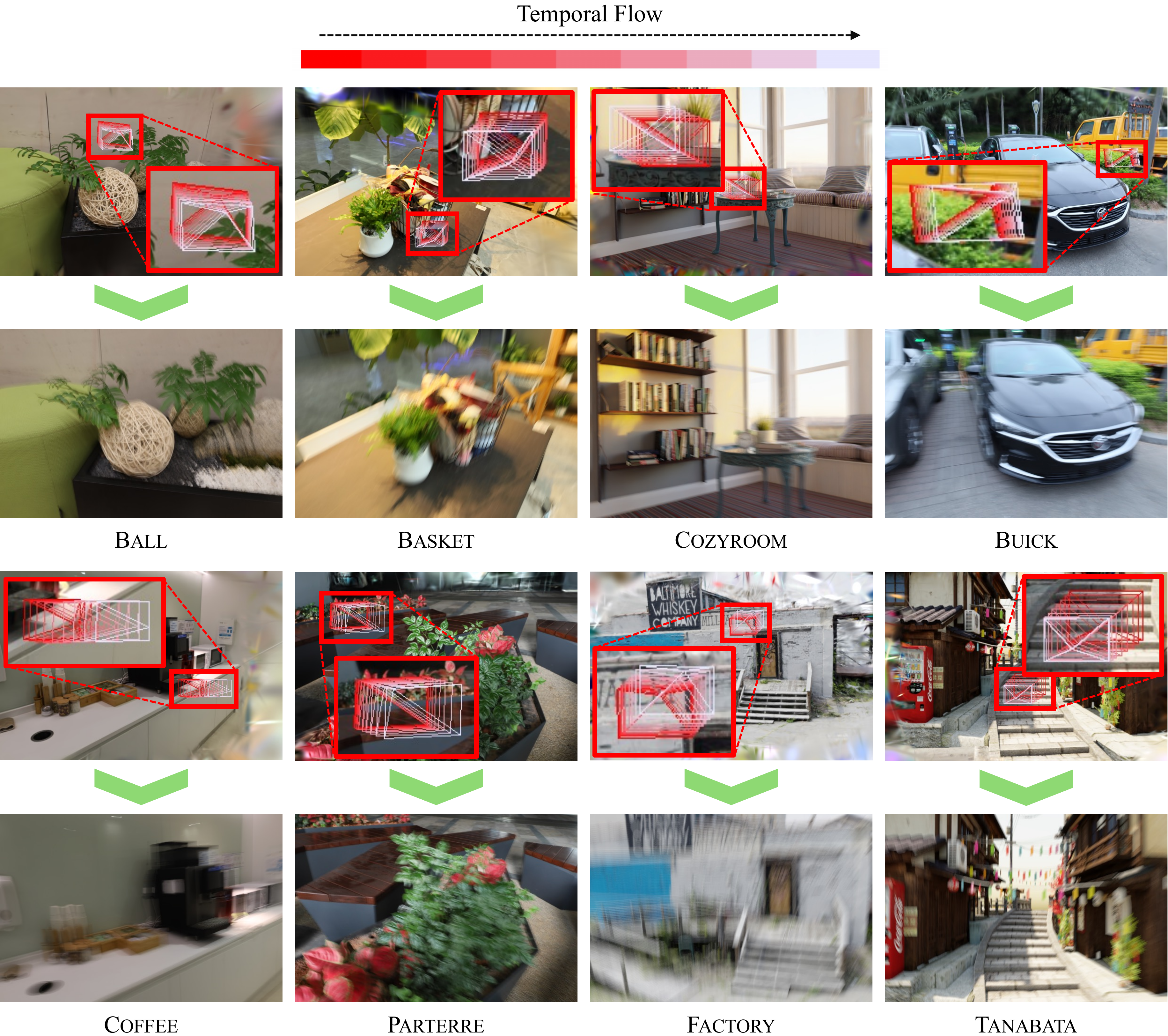}
	\caption{Camera motion trajectory predicted by CRiM-GS for input images with significant blur.}
	\label{fig:kernel_visualize_extreme}
\end{figure*}

\begin{figure*}[t]
	\centering
	\includegraphics[width=0.9\linewidth]{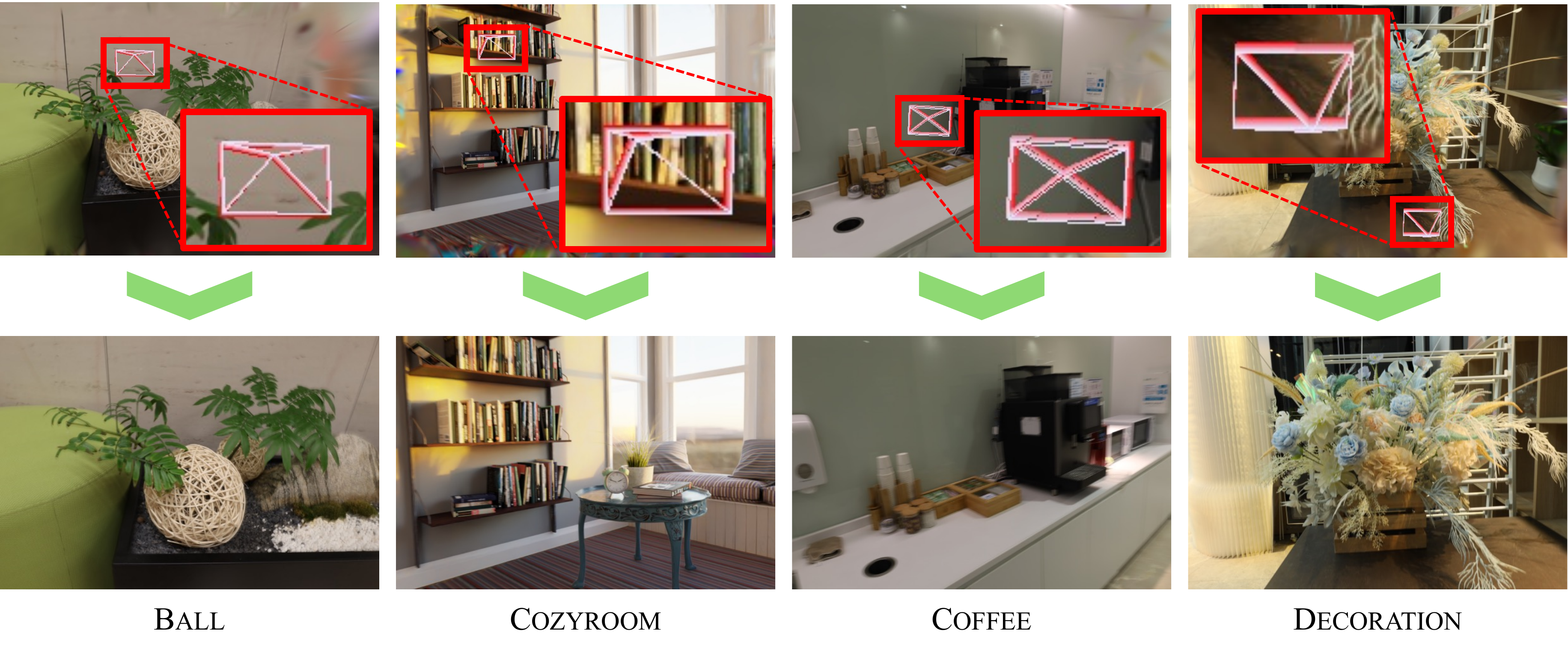}
	\caption{Camera motion trajectory predicted by CRiM-GS for input images with moderate blur.}
	\label{fig:kernel_visualize_moderate}
\end{figure*}

\paragraph{Number of Poses on Camera Motion}
We conduct ablation experiments on the number of camera poses composing the continuous camera motion, and the results are shown in \cref{tab:num_pose}. The results indicate that performance improves across all metrics as $N$ increases, suggesting that a larger number of camera poses allows for more precise modeling of the camera motion trajectory. However, beyond $N=8$, the performance stabilizes and based on these observations, we adopt $N=9$ as it offers the best overall performance.

\begin{table}[t]
	\caption{Experimental results based on the number of poses $N$ along the camera motion trajectory.}
	\vspace{-3mm}
	\begin{center}
		\resizebox{\columnwidth}{!}{
			\centering
			\setlength{\tabcolsep}{2pt}
			\begin{tabular}{l||c|c|c|c|c|c}
				\toprule 
				
				\multirow{2}{*}{Methods} 			  	& \multicolumn{3}{c|}{Synthetic Scene Dataset}  	   & \multicolumn{3}{c}{Real-World Scene Dataset}  	\\ \cmidrule{2-7}
				&  ~PSNR$\uparrow$~    &~SSIM$\uparrow$~    &~LPIPS$\downarrow$~ &~PSNR$\uparrow$~    &~SSIM$\uparrow$~    &~LPIPS($\downarrow$)~     	\\ \midrule \midrule
				$N$=5			& 27.40	& 0.8314	& 0.0936	& 26.42	& 0.8040	& 0.0924 \\
				$N$=6		& 28.78	& 0.8610	& 0.711	& 26.77	& 0.8082	& 0.0822\\
				$N$=7		& 29.52	& 0.8758	& 0.0655	& 27.01 	& 0.8145	& 0.0762\\
				$N$=8 	& 30.17 & 0.9002	& 0.0501	& 26.81	& 0.8101	& 0.0699 \\ 
				$N$=9 	& \cellcolor{second!35}30.44	& \cellcolor{second!35}0.9095	& \cellcolor{best!25}0.0441	& \cellcolor{best!25}27.33	& \cellcolor{best!25}0.8310	& \cellcolor{second!35}0.0634 \\ 
				$N$=10	& \cellcolor{best!25}30.58	& \cellcolor{best!25}0.9124	& \cellcolor{second!35}0.0445	& \cellcolor{second!35}27.12	& \cellcolor{second!35}0.8227	& \cellcolor{best!25}0.0630 \\ \bottomrule
			\end{tabular}
		}
	\end{center}
	\label{tab:num_pose}
	\vspace{-5mm}
\end{table}

\section{Difference from SMURF~\cite{lee2024smurf}}
In this section, we compare our approach with SMURF, a methodology for handling the continuous dynamics of camera motion blur. SMURF utilizes neural ODE to warp a given input ray into continuous rays that simulate camera motion. However, its continuous dynamics are applied only in the 2D pixel space, lacking the inclusion of higher-dimensional camera motion in 3D space. Additionally, as SMURF is implemented on Tensorial Radiance Fields (TensoRF)~\cite{chen2022tensorf}, a ray tracing-based method, it exhibits relatively slower training and rendering speeds.

In contrast, our model uses neural ODE to obtain the 3D camera poses which constitue the camera motion trajectory. Our approach incorporates higher-dimensional information compared to SMURF by operating directly in 3D space rather than the 2D pixel space. Furthermore, as our method is implemented on 3DGS~\cite{kerbl20233d}, a rasterization-based method, it ensures faster training and rendering speeds than SMURF.

\section{Post-Training Pose Optimization}
We reproduce the results of DeblurGS~\cite{oh2024deblurgs} as it performs post-training optimization of test camera poses. This process requires test images and input test camera poses to obtain optimal poses, making it difficult to consider a fair comparison. Nevertheless, as shown in \cref{tab:pose}, CRiM-GS achieves better performance with the same pose optimization process as DeblurGS, and even without the post-training optimization, CRiM-GS outperforms the DeblurGS with optimized poses. This indicates that CRiM-GS is robust to inaccurate camera poses. As illustrated in the error maps in \cref{fig:pose}, CRiM-GS not only performs better than DeblurGS but also optimizes poses better than other models without additional training.

\begin{table}[t]
	\caption{Experimental results based on whether pose optimization is performed after training}
	\begin{center}
		\resizebox{\columnwidth}{!}{
			\centering
			\setlength{\tabcolsep}{2pt}
			\begin{tabular}{l||c|c|c|c|c|c|c}
				\toprule 
				
				\multirow{2}{*}{Methods} 			  & \multirow{2}{*}{~P.O.~} 	& \multicolumn{3}{c|}{Synthetic Scene Dataset}  	   & \multicolumn{3}{c}{Real-World Scene Dataset}  	\\ \cmidrule{3-8}
				&  &~PSNR$\uparrow$~    &~SSIM$\uparrow$~    &~LPIPS$\downarrow$~ &~PSNR$\uparrow$~    &~SSIM$\uparrow$~    &~LPIPS($\downarrow$)~     	\\ \midrule \midrule
				
				DeblurGS*~\cite{oh2024deblurgs}					&				& 20.22		& 0.5454	& 0.1042			& 20.48		& 0.5813		& 0.1186 \\
				\textbf{CRiM-GS}											    &                    & \cellcolor{best!25}30.44    & \cellcolor{best!25}0.9095   & \cellcolor{best!25}0.0441 		 & \cellcolor{best!25}27.33       & \cellcolor{best!25}0.8310      & \cellcolor{best!25}0.0634	\\ \midrule
				DeblurGS*~\cite{oh2024deblurgs}					&\ding{51}				& 30.24		& 0.8922	& 0.0637			& 26.07		& 0.8024		& 0.0958 \\			
				\textbf{CRiM-GS}											          &   \ding{51}           & \cellcolor{best!25}30.65    & \cellcolor{best!25}0.9123   & \cellcolor{best!25}0.0435 		 & \cellcolor{best!25}28.34       & \cellcolor{best!25}0.8478      & \cellcolor{best!25}0.0606 	\\ \bottomrule
			\end{tabular}
		}
	\end{center}
	\label{tab:pose}
\end{table}

\section{Blur Kernel Visualization}
We visualize the camera motion trajectories for input blurry images predicted by CRiM-GS in \cref{fig:kernel_visualize_extreme} and \cref{fig:kernel_visualize_moderate}. \cref{fig:kernel_visualize_extreme} illustrates camera motions for images with significant blur, where the predicted trajectories are continuous over time and align precisely with the input images. \cref{fig:kernel_visualize_moderate} depicts camera motions for images with relatively less blur, where the predicted trajectories show minimal movement, yet still match the input images accurately. These results demonstrate that our blurring kernel effectively models precise continuous camera motion.

\section{Derivation of Rigid Body Motion~\cite{lynch2017modernrobotics}}
In this section, we explain the derivation process for Eq.~(\textcolor{red}{9}) and Eq.~(\textcolor{red}{10}) from the main paper. This derivation aims to expand and simplify the process described in Modern Robotics~\cite{lynch2017modernrobotics} for better clarity and accessibility.

The components of a given screw axis include the unit rotation axis $\hat{\omega}\in\mathbb{R}^{3}$ and the translation component $v\in\mathbb{R}^{3}$. The unit rotation axis consists of the angular velocity $\omega$ and the rotation angle $\theta$:
\begin{equation}
	\hat{\omega}=\frac{\omega}{\theta},\quad where\quad||\hat{\omega}||=1.
\end{equation}
We combine the rotation axis $\hat{\omega}$ and the rotation angle $\theta$ to represent an element of the Lie Algebra, $\mathfrak{so}(3)$, which serves as the linear approximation of the rotation matrix. Before proceeding, $\hat{\omega}$ is converted into a $3\times 3$ skew-symmetric matrix $[\hat{\omega}]$ to compactly express the cross-product operation as a matrix multiplication:
\begin{equation}\label{eq:skew}
	[\hat{\omega}]=\left[
		\begin{array}{ccc}
			0 & -\hat{\omega}_{z} & \hat{\omega}_{y} \\
			\hat{\omega}_{z} & 0 & -\hat{\omega}_{x} \\
			-\hat{\omega}_{y} & \hat{\omega}_{x} & 0 \\
		\end{array}	
	\right] 
	\in \mathfrak{so}(3),~where~ [\hat{\omega}]^3 = I.
\end{equation}
Using the skew-symmetric matrix $[\hat{\omega}]$ and the translation component $v$, the screw axis $[S]$ is expressed. By multiplying this screw axis with $\theta$, we incorporate the magnitude of the rotation and translation along the screw axis:
\begin{equation}
	[\mathcal{S}]\theta=\left[
		\begin{array}{cc}
			[\hat{\omega}]\theta & v\theta \\
			0 & 0 \\
		\end{array}
	\right]
	\in \mathfrak{se}(3),
\end{equation}
where$\mathfrak{se}(3)$ represents the Lie Algebra, which corresponds to the infinitesimal changes of the Lie Group \textit{SE}(3). To map this infinitesimal change to the \textit{SE}(3) transformation matrix $\mathbf{T}=e^{[\mathcal{S}]\theta}$, we use the Taylor expansion, following these steps:\\
{\small{\begin{align}
	e^{[\mathcal{S}]\theta} & = \sum_{n=0}^{\infty} [\mathcal{S}]^{n}\frac{\theta^{n}}{n!} \\
											  & = I + [\mathcal{S}]\theta + [\mathcal{S}]^2\frac{\theta^2}{2!} + \cdots \\
											  	& = \left[\begin{array}{cc}
											  		I + [\hat{\omega}]\theta + [\hat{\omega}]^{2}\frac{\theta^{2}}{2!}+\cdots & \left(I\theta + [\hat{\omega}]\frac{\theta^{2}}{2!}+\cdots\right)v \\
											  		0 & 1 \\
											  	\end{array}
											  	\right] \\
											  	& = \left[\begin{array}{cc}
											  		e^{[\hat{\omega}]\theta} & G(\theta)v \\
											  		0 & 1 \\
											  		\end{array}
											  	\right] \in \textit{SE}(3)\\
											  	& \because~[\mathcal{S}]^{n}=\left[
											  	\begin{array}{cc}
											  		[\hat{\omega}]^{n} & [\hat{\omega}]^{n-1}v \\
											  		0 & 0 \\
											  	\end{array}
											  	\right]
\end{align}}}
For the rotation matrix $e^{[\hat{\omega}]}\theta$, we simplify it using the Taylor expansion and \cref{eq:skew}, resulting in:
{\small{\begin{align}
	e^{[\hat{\omega}]\theta} & = I + [\hat{\omega}]\theta + [\hat{\omega}]^{2}\frac{\theta^{2}}{2!} + [\hat{\omega}]^{3}\frac{\theta^{3}}{3!} +  + [\hat{\omega}]^{4}\frac{\theta^{4}}{4!}\cdots \\
											  & = I + \left(\theta - \frac{\theta^{3}}{3!} + \cdots\right)[\hat{\omega}] + \left(\frac{\theta^{2}}{2!} - \frac{\theta^{4}}{4!} + \cdots\right)[\hat{\omega}]^{2} \\
											  & = I + \sin \theta[\hat{\omega}] + (1 - \cos \theta)[\hat{\omega}]^2 \in \textit{SO}(3)
\end{align}}}
The translational component $G(\theta)$ is also derived using the Taylor expansion and \cref{eq:skew}:
{\small{\begin{align}
	G(\theta) & = I\theta + [\hat{\omega}]\frac{\theta^{2}}{2!} + [\hat{\omega}]^{2}\frac{\theta^{3}}{3!} + [\hat{\omega}]^{3}\frac{\theta^{4}}{4!}+\cdots \\
	& = I\theta + \left(\frac{\theta^{2}}{2!} - \frac{\theta^{4}}{4!} + \cdots\right)[\hat{\omega}] + \left(\frac{\theta^{3}}{3!} - \frac{\theta^{5}}{5!} + \cdots\right)[\hat{\omega}]^{2} \\
	& = I\theta + (1 - \cos \theta)[\hat{\omega}] + (\theta - \sin \theta)[\hat{\omega}]^2
\end{align}}}
The term $G(\theta)$ physically represents the total translational motion caused by the rotational motion as the rigid body rotates by $\theta$. In other words, $G(\theta)$ indicates how rotational motion contributes to translational motion, which can also be expressed as an integral of the rotation motion:
\begin{align}
	G(\theta) & = \int_{0}^{\theta} e^{[\hat{\omega}]\theta}d\theta \\
	& = \int_{0}^{\theta} \left(I + \sin \theta[\hat{\omega}] + (1 - \cos \theta)[\hat{\omega}]^2\right)d\theta \\
	& = I\theta + (1 - \cos \theta)[\hat{\omega}] + (\theta - \sin \theta)[\hat{\omega}]^2
\end{align}
Through the above process, we derive Eq.~(\textcolor{red}{9}) and Eq.~(\textcolor{red}{10}) in the main paper, improving readability of the paper and providing a clear foundation for understanding the mathematical framework.

\section{Per-Scene Quantitative Results}
We show the per-scene quantitative performance on synthetic and real-world datasets in \cref{tab:synthetic} and \cref{tab:real}. CRiM-GS demonstrates superior performance on most scenes in the real-world dataset and achieves the highest performance across all scenes in the synthetic dataset. Notably, CRiM-GS achieves the best LPIPS scores for all scenes, highlighting the superior quality achieved by our CRiM-GS.

\section{Additional Qualitative Results}
We provide additional visualization results in \cref{fig:additional}, which demonstrate that our CRiM-GS outperforms not only in quantitative metrics but also in qualitative performance. For comparative videos, please refer to \textit{the supplementary materials}.

\paragraph{Error Map Visualization.}
We extract error maps for three scenes to enable direct comparison with other methods, and the results are shown in \cref{fig:pose}. While BAD-GS~\cite{zhao2024bad} and DeblurGS~\cite{oh2024deblurgs} exhibit relatively high errors, BAGS~\cite{peng2024bags} and CRiM-GS demonstrate lower errors. Notably, our model captures the overall contours of the scene more effectively than BAGS.

\section{Limitation}
Despite achieving superior qualitative and quantitative performance with fast rendering, the proposed model faces room for optimization in terms of training efficiency. The primary challenge lies in the multiple rendering steps per iteration and the use of the CNN module $\mathcal{F}$ for pixel-wise weighted sums, which is employed only during training. However, this aspect offers opportunities for future improvements. Designing continuous motion directly at the 3D Gaussian level could streamline the process, reducing training time without compromising performance.


\begin{table*}[t]
	\caption{Per-Scene Quantitative Performance on the Real-World Scenes.}
	\begin{center}
		\resizebox{\linewidth}{!}{
			\centering
			\setlength{\tabcolsep}{1.5pt}
			\scriptsize
			\begin{tabular}{l||c|c|c|c|c|c|c|c|c|c|c|c|c|c|c}
				\toprule 
				
				\multirow{2}{*}{Real-World Scene\ } 			   & \multicolumn{3}{c|}{\textsc{Ball}}  	   & \multicolumn{3}{c|}{\textsc{Basket}}  	   & \multicolumn{3}{c|}{\textsc{Buick}}  				& \multicolumn{3}{c|}{\textsc{Coffee}} 		 & \multicolumn{3}{c}{\textsc{Decoration}}  \\ \cmidrule{2-16}
				&\ PSNR$\uparrow$ \     &\ SSIM$\uparrow$ \     &\ LPIPS$\downarrow$ \  &\ PSNR$\uparrow$ \     &\ SSIM$\uparrow$ \     &\ LPIPS$\downarrow$ \ 		&\ PSNR$\uparrow$ \     &\ SSIM$\uparrow$ \     &\ LPIPS$\downarrow$ \ 		&\ PSNR$\uparrow$ \     &\ SSIM$\uparrow$ \       &\ LPIPS$\downarrow$ \ 	&\ PSNR$\uparrow$ \     &\ SSIM$\uparrow$ \       &\ LPIPS$\downarrow$ \   	\\ \midrule \midrule
				Naive NeRF~\cite{mildenhall2020nerf}                            				 & 24.08    & 0.6237   & 0.3992 		& 23.72       & 0.7086      & 0.3223      & 21.59    & 0.6325      & 0.3502    		& 26.48    & 0.8064      & 0.2896    		& 22.39    & 0.6609     & 0.3633    \\ \midrule
				Deblur-NeRF~\cite{ma2022deblurnerf}                         				& 27.36    & 0.7656   & 0.2230 		 & 27.67       & 0.8449      & 0.1481        & 24.77     & 0.7700      & 0.1752    		& 30.93     & 0.8981      & 0.1244    		 & 24.19    & 0.7707     & 0.1862    \\
				PDRF-10*~\cite{peng2023pdrf}                            				 & 27.37    & 0.7642   & 0.2093 		& 28.36       & 0.8736      & 0.1179      & 25.73    & 0.7916      & 0.1582    		& \cellcolor{second!35}31.79    & 0.9002      & 0.1133    		& 23.55   & 0.7508     & 0.2145    \\ 
				BAD-NeRF*~\cite{wang2023bad}                            				 & 21.33 & 0.5096 & 0.4692 			& 26.44 		& 0.8080 	& 0.1325 		& 21.63  & 0.6429 		& 0.2593 		& 28.98 & 0.8369 		& 0.1956 		& 22.13	 	& 0.6316 	& 0.2894    \\ 
				DP-NeRF~\cite{lee2023dp}                        						  & 27.20    & 0.7652   & 0.2088 		& 27.74   	 & 0.8455      & 0.1294    		& 25.70    & 0.7922      & 0.1405    	& 31.19    & 0.9049      & 0.1002    		& 24.31    & 0.7811     & 0.1639   \\ \midrule
				DeblurGS~\cite{oh2024deblurgs}											& 21.44		& 0.5760		& 0.1480		& 21.00		& 0.6706		& 0.0908			& 18.91		& 0.5539		& 0.1442	& 25.97	& 0.8299		& \cellcolor{second!35}0.0614		& 19.60	& 0.5713		& 0.1035 \\
				BAD-GS*~\cite{zhao2024bad}                            				 & 22.14    & 0.5791   & \cellcolor{second!35}0.1095 		& 21.48       & 0.6787      & \cellcolor{second!35}0.0603      & 18.81    & 0.5271      & 0.0952    		& 24.54    & 0.7627      & 0.1031    		& 20.30   & 0.6162     & 0.0881    \\ 
				Deblurring 3DGS~\cite{lee2024deblurring}												& 28.27 & 0.8233 & 0.1413 & 28.42 & 0.8713 & 0.1155		& 25.95		& 0.8367		& 0.0954		& 32.84		& 0.9312 & 	0.0676	& 25.87 & 0.8540 &  0.0933 \\
				BAGS~\cite{peng2024bags}                            				 & \cellcolor{second!35}27.68    & \cellcolor{second!35}0.7990   & 0.1500 		& \cellcolor{second!35}29.54       & \cellcolor{second!35}0.9000      & 0.0680      & \cellcolor{second!35}26.18    & \cellcolor{best!25}0.8440      & \cellcolor{second!35}0.0880    		& 31.59    & \cellcolor{second!35}0.9080      & 0.0960    		& \cellcolor{second!35}26.09   & \cellcolor{second!35}0.8580     & \cellcolor{second!35}0.0830    \\ \midrule
				\textbf{CRiM-GS}                            				 & \cellcolor{best!25}28.25    & \cellcolor{best!25}0.8064   & \cellcolor{best!25}0.1074 		& \cellcolor{best!25}30.85       & \cellcolor{best!25}0.9048      & \cellcolor{best!25}0.0425      & \cellcolor{best!25}26.51    & \cellcolor{second!35}0.8379      & \cellcolor{best!25}0.0585    		& \cellcolor{best!25}32.26    & \cellcolor{best!25}0.9230      & \cellcolor{best!25}0.0421    		& \cellcolor{best!25}26.34   & \cellcolor{best!25}0.8610     & \cellcolor{best!25}0.0569    \\ \midrule \midrule
				
				\multirow{2}{*}{Real-World Scene\ } 			   & \multicolumn{3}{c|}{\textsc{Girl}}  	   & \multicolumn{3}{c|}{\textsc{Heron}}  	   & \multicolumn{3}{c|}{\textsc{Parterre}}  				& \multicolumn{3}{c|}{\textsc{Puppet}} 		 & \multicolumn{3}{c}{\textsc{Stair}}  \\ \cmidrule{2-16}
				&\ PSNR$\uparrow$ \     &\ SSIM$\uparrow$ \     &\ LPIPS$\downarrow$ \  &\ PSNR$\uparrow$ \     &\ SSIM$\uparrow$ \     &\ LPIPS$\downarrow$ \ 		&\ PSNR$\uparrow$ \     &\ SSIM$\uparrow$ \     &\ LPIPS$\downarrow$ \ 		&\ PSNR$\uparrow$ \     &\ SSIM$\uparrow$ \       &\ LPIPS$\downarrow$ \ 	&\ PSNR$\uparrow$ \     &\ SSIM$\uparrow$ \       &\ LPIPS$\downarrow$ \   	\\ \midrule \midrule
				Naive NeRF~\cite{mildenhall2020nerf}                            				 & 20.07    & 0.7075   & 0.3196 		& 20.50       & 0.5217      & 0.4129      & 23.14    & 0.6201      & 0.4046    		& 22.09    & 0.6093      & 0.3389    		& 22.87    & 0.4561     & 0.4868    \\ \midrule
				Deblur-NeRF~\cite{ma2022deblurnerf}                         				& 22.27    & 0.7976   & 0.1687 		 & 22.63       & 0.6874      & 0.2099        & 25.82     & 0.7597      & 0.2161    		& 25.24     & 0.7510      & 0.1577    		 & 25.39    & 0.6296     & 0.2102    \\
				PDRF-10*~\cite{peng2023pdrf}                            				 & 24.12    & 0.8328   & 0.1679 		& 22.53       & 0.6880      & 0.2358      & 25.36    & 0.7601      & 0.2263    		& 25.02    & 0.7496     & 0.1532    		& 25.20    & 0.6235     & 0.2288    \\ 
				BAD-NeRF*~\cite{wang2023bad}                            				 & 18.10 & 0.5652 	& 0.3933 		& 22.18 		& 0.6479 	& 0.2226 		& 23.44 	& 0.6243 	& 0.3151 		& 22.48 	& 0.6249 	& 0.2762 			& 21.52 	& 0.4237 	& 0.3341    \\ 
				DP-NeRF~\cite{lee2023dp}                        						  & 23.33    & 0.8139   & 0.1498 		& \cellcolor{best!25}22.88   	 & 0.6930      & 0.1914    		& 25.86    & 0.7665      & 0.1900    	& 25.25    & 0.7536      & 0.1505    		& 25.59    & 0.6349      & 0.1772   \\ \midrule 
				DeblurGS*~\cite{oh2024deblurgs}												& 18.28		& 0.6660	& 0.1044		& 18.70	& 0.4579	& 0.1488		& 19.78		& 0.4953	& 0.1744	& 19.50	& 0.5302	& 0.1297		& 21.63	& 0.4623		& 0.0808	\\
				BAD-GS*~\cite{zhao2024bad}                            				 	& 18.72    & 0.6765   & 0.0925 		& 18.53       & 0.4498      & 0.1412      & 20.79    & 0.5589      & 0.1298   		& 21.08    & 0.6036     & 0.1022    		& 21.77    & 0.4640     & \cellcolor{second!35}0.0782    \\ 
				Deblurring 3DGS~\cite{lee2024deblurring}							& 23.26 & 0.8390 & 	0.1011	& 23.14 & 0.7438 & 0.1543	 		& 26.17 & 0.8144	& 0.1206		& 25.67 & 0.8051	& 0.0941		& 26.46 & 0.7050 & 0.1123 \\
				BAGS~\cite{peng2024bags}                            				 		& \cellcolor{second!35}25.45    & \cellcolor{second!35}0.8690   & \cellcolor{second!35}0.0790 		& 22.04       & \cellcolor{second!35}0.7150      & \cellcolor{second!35}0.1260      & \cellcolor{second!35}25.92    & \cellcolor{best!25}0.8190      & \cellcolor{second!35}0.0920    		& \cellcolor{second!35}25.81    & \cellcolor{second!35}0.8040     & \cellcolor{second!35}0.0940    		& \cellcolor{second!35}26.69    & \cellcolor{second!35}0.7210     & 0.0800    \\ \midrule
				\textbf{CRiM-GS}                            				 	& \cellcolor{best!25}26.20    & \cellcolor{best!25}0.8780   & \cellcolor{best!25}0.0438 		& \cellcolor{second!35}22.84       & \cellcolor{best!25}0.7234      & \cellcolor{best!25}0.1110     & \cellcolor{best!25}26.09    & \cellcolor{second!35}0.8126      & \cellcolor{best!25}0.0698    		& \cellcolor{best!25}26.66    & \cellcolor{best!25}0.8236     & \cellcolor{best!25}0.0613    		& \cellcolor{best!25}27.30    & \cellcolor{best!25}0.7390     & \cellcolor{best!25}0.0415    \\ \bottomrule
			\end{tabular}
		}
	\end{center}
	\label{tab:real}
\end{table*}

\begin{table*}[t]
	\caption{Per-Scene Quantitative Performance on the Synthetic Scenes.}
	\begin{center}
		\resizebox{\linewidth}{!}{
			\centering
			\setlength{\tabcolsep}{1.5pt}
			\scriptsize
			\begin{tabular}{l||c|c|c|c|c|c|c|c|c|c|c|c|c|c|c}
				\toprule 
				
				\multirow{2}{*}{Synthetic Scene} 			   & \multicolumn{3}{c|}{\textsc{Factory}}  	   & \multicolumn{3}{c|}{\textsc{CozyRoom}}  	   & \multicolumn{3}{c|}{\textsc{Pool}}  				& \multicolumn{3}{c|}{\textsc{Tanabata}} 		 & \multicolumn{3}{c}{\textsc{Trolley}}  \\ \cmidrule{2-16}
				&\ PSNR$\uparrow$ \     &\ SSIM$\uparrow$ \     &\ LPIPS$\downarrow$ \  &\ PSNR$\uparrow$ \     &\ SSIM$\uparrow$ \     &\ LPIPS$\downarrow$ \ 		&\ PSNR$\uparrow$ \     &\ SSIM$\uparrow$ \     &\ LPIPS$\downarrow$ \ 		&\ PSNR$\uparrow$ \     &\ SSIM$\uparrow$ \       &\ LPIPS$\downarrow$ \ 	&\ PSNR$\uparrow$ \     &\ SSIM$\uparrow$ \       &\ LPIPS$\downarrow$ \   	\\ \midrule \midrule
				Deblur-NeRF~\cite{ma2022deblurnerf}                         				& 25.60    & 0.7750   & 0.2687 		 & 32.08       & 0.9261      & 0.0477        & 31.61     & 0.8682      & 0.1246    		& 27.11     & 0.8640      & 0.1228    		 & 27.45    & 0.8632     & 0.1363    \\
				PDRF-10*~\cite{peng2023pdrf}                         					   & 25.87    & \cellcolor{second!35}0.8316   & 0.1915 		 		  & 31.13       &  0.9225     & 0.0439        			& 31.00     & 0.8583      & 0.1408    				& \cellcolor{second!35}28.01     & \cellcolor{second!35}0.8931      & 0.1004    		 		 & \cellcolor{second!35}28.29    & \cellcolor{second!35}0.8921     & 0.0931 		    \\
				BAD-NeRF*~\cite{wang2023bad}                         				& 24.43    & 0.7274   & 0.2134 		 & 29.77       & 0.8864      & 0.0616        & 31.51     & 0.8620      & 0.0802    		& 25.32     & 0.8081      & 0.1077    		 & 25.58    & 0.8049     & 0.1008    \\
				DP-NeRF~\cite{lee2023dp}                        				 	 & \cellcolor{second!35}25.91    & 0.7787   & 0.2494 		& \cellcolor{second!35}32.65   	 & 0.9317      & 0.0355    		& \cellcolor{second!35}31.96    & \cellcolor{second!35}0.8768      & 0.0908    	& 27.61    & 0.8748      & 0.1033    		& 28.03    & 0.8752      & 0.1129   \\ \midrule
				DeblurGS*~\cite{oh2024deblurgs}											& 17.82		& 0.3980	& \cellcolor{second!35}0.1453		& 22.88		& 0.6758	& 0.0514		& 24.53		& 0.6095	& 0.1171		& 17.54		& 0.4800		& 0.1132		& 18.32		& 0.5638		& 0.0942 \\
				BAD-GS*~\cite{zhao2024bad}										 	 & 16.84    & 0.3220   & 0.2425 			& 22.31       & 0.6954      & 0.0519      & 26.70    & 0.7051      & \cellcolor{second!35}0.0642    		& 20.13    & 0.6095      & \cellcolor{second!35}0.0729    		& 19.60    & 0.5941     & \cellcolor{second!35}0.0722    \\ 
				Deblurring 3DGS~\cite{lee2024deblurring}												& 24.01    & 0.7333   & 0.2326 			& 31.45       & 0.9222      & 0.0367      & 31.87    & 0.8829      & 0.0751    		& 27.01    & 0.8807      & 0.0785    		& 26.88    & 0.8710     & 0.1028    \\
				BAGS~\cite{peng2024bags}												& 22.35    & 0.6639   & 0.2277 			& 32.21       & \cellcolor{second!35}0.9359      & \cellcolor{second!35}0.0245      & 28.72    & 0.8404      & 0.0804    		& 26.79    & 0.8735      & 0.1099    		& 26.61    & 0.8627     & 0.1156    \\ \midrule \midrule
				\textbf{CRiM-GS}											 & \cellcolor{best!25}28.72    & \cellcolor{best!25}0.8874   & \cellcolor{best!25}0.0604 			& \cellcolor{best!25}33.01       & \cellcolor{best!25}0.9409      & \cellcolor{best!25}0.0209      & \cellcolor{best!25}32.05    & \cellcolor{best!25}0.8882      & \cellcolor{best!25}0.0524    		& \cellcolor{best!25}29.04    & \cellcolor{best!25}0.9180      & \cellcolor{best!25}0.0376    		& \cellcolor{best!25}29.36    & \cellcolor{best!25}0.9131     & \cellcolor{best!25}0.0492    \\  \bottomrule
			\end{tabular}
		}
		\label{tab:synthetic}
	\end{center}
	\vspace{-5mm}
\end{table*}

\begin{table*}[t]
	\begin{center}
		\setlength\tabcolsep{1pt}
		\begin{tabular}{ccccc}
\small{Reference} & \footnotesize{DP-NeRF~\cite{oh2024deblurgs}}  & \footnotesize{BAGS~\cite{peng2024bags}}  & \footnotesize{CRiM-GS (Ours)}  & \footnotesize{Ground Truth} \\
			\includegraphics[width=0.19\textwidth]{for_figa/buick_reference.jpg} &			
			\includegraphics[width=0.19\textwidth]{for_fig3/buick_2_dpnerf.png} &
			\includegraphics[width=0.19\textwidth]{for_fig3/buick_2_bags.png} &
			\includegraphics[width=0.19\textwidth]{for_fig3/buick_2_ours.png} &
			\includegraphics[width=0.19\textwidth]{for_fig3/buick_2_gt.png} \\
			
			\includegraphics[width=0.19\textwidth]{for_figa/puppet_reference.jpg} &			
			\includegraphics[width=0.19\textwidth]{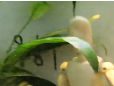} &
			\includegraphics[width=0.19\textwidth]{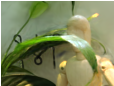} &
			\includegraphics[width=0.19\textwidth]{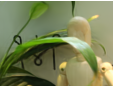} &
			\includegraphics[width=0.19\textwidth]{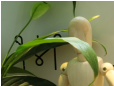} \\
			
			\includegraphics[width=0.19\textwidth]{for_figa/tanabata_reference.jpg} &
			\includegraphics[width=0.19\textwidth]{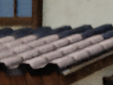} &
			\includegraphics[width=0.19\textwidth]{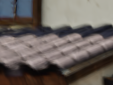} &
			\includegraphics[width=0.19\textwidth]{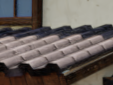} &
			\includegraphics[width=0.19\textwidth]{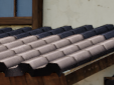} \\

			\includegraphics[width=0.19\textwidth]{for_figa/parterre_reference.jpg} &			
			\includegraphics[width=0.19\textwidth]{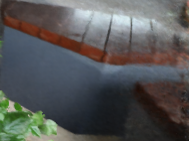} &
			\includegraphics[width=0.19\textwidth]{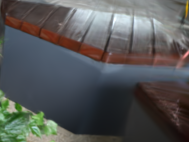} &
			\includegraphics[width=0.19\textwidth]{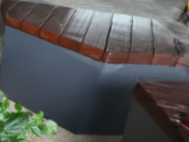} &
			\includegraphics[width=0.19\textwidth]{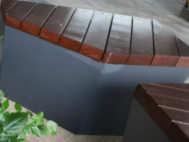} \\

			\includegraphics[width=0.19\textwidth]{for_figa/wine_reference.jpg} &			
			\includegraphics[width=0.19\textwidth]{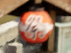} &
			\includegraphics[width=0.19\textwidth]{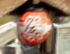} &
			\includegraphics[width=0.19\textwidth]{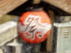} &
			\includegraphics[width=0.19\textwidth]{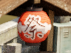} \\
			
			\includegraphics[width=0.19\textwidth]{for_figa/basket_reference.jpg} &			
			\includegraphics[width=0.19\textwidth]{for_figa/basket_dpnerf.png} &
			\includegraphics[width=0.19\textwidth]{for_figa/basket_bags.png} &
			\includegraphics[width=0.19\textwidth]{for_figa/basket_ours.png} &
			\includegraphics[width=0.19\textwidth]{for_figa/basket_gt.png} \\
			
			\includegraphics[width=0.19\textwidth]{for_figa/stair_reference.jpg} &			
			\includegraphics[width=0.19\textwidth]{for_figa/stair_dpnerf.png} &
			\includegraphics[width=0.19\textwidth]{for_figa/stair_bags.png} &
			\includegraphics[width=0.19\textwidth]{for_figa/stair_ours.png} &
			\includegraphics[width=0.19\textwidth]{for_figa/stair_gt.png} \\
			
		\end{tabular}
		\captionof{figure}{Additional Qualitative Comparison on the Synthetic and Real-World Scenes.}
		\label{fig:additional}
	\end{center}
\end{table*}

%
%
%
%

\begin{table*}[t]
	\begin{center}
		\setlength\tabcolsep{1pt}
		\begin{tabular}{cccc}
			\footnotesize{BAD-GS~\cite{zhao2024bad}} & \footnotesize{DeblurGS~\cite{oh2024deblurgs}}  & \footnotesize{BAGS~\cite{peng2024bags}}  & \footnotesize{CRiM-GS (Ours)} \\
			\includegraphics[width=0.24\textwidth]{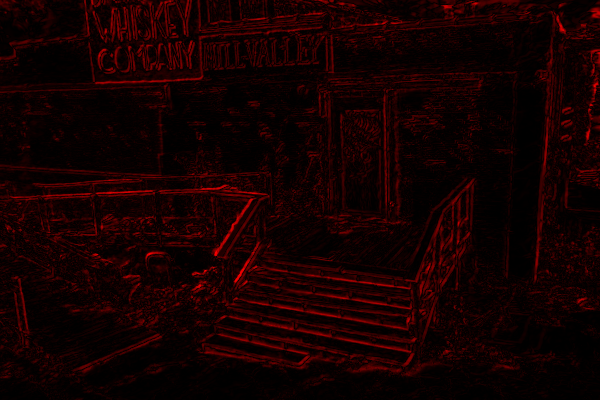} &
			\includegraphics[width=0.24\textwidth]{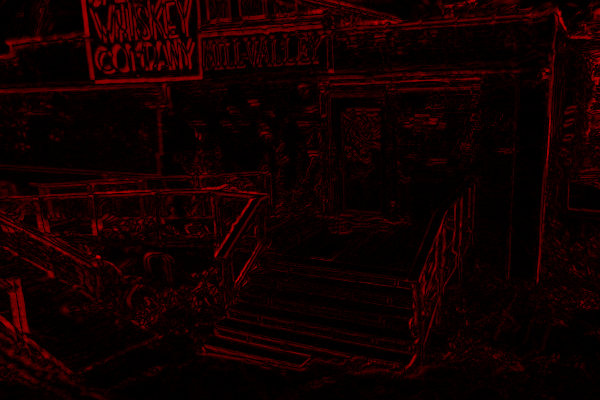} &
			\includegraphics[width=0.24\textwidth]{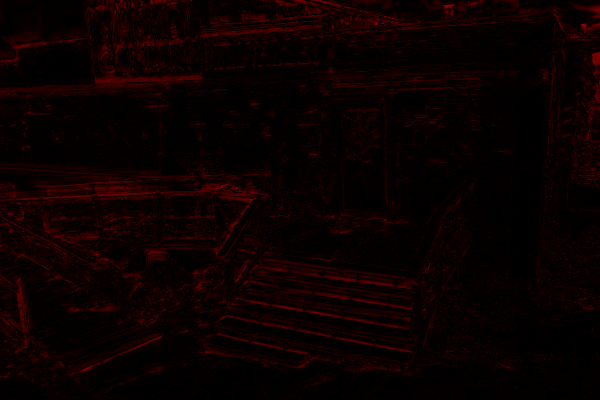} &
			\includegraphics[width=0.24\textwidth]{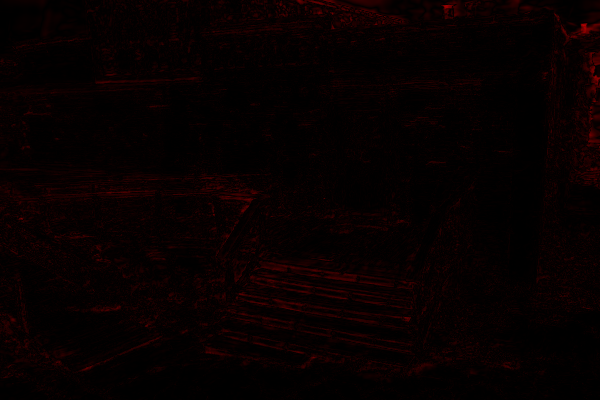} \\
			
			\includegraphics[width=0.24\textwidth]{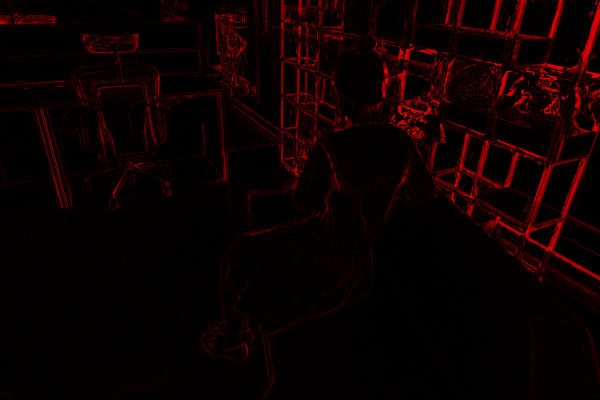} &
			\includegraphics[width=0.24\textwidth]{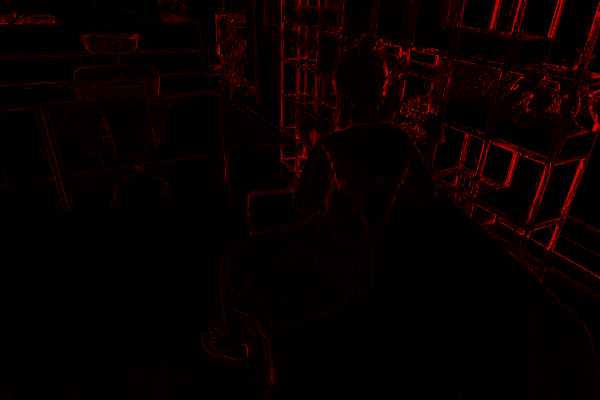} &
			\includegraphics[width=0.24\textwidth]{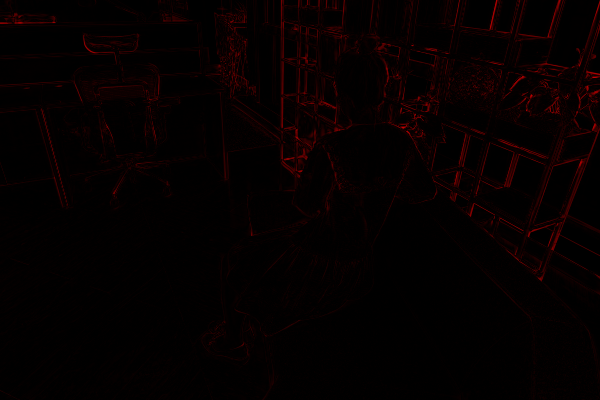} &
			\includegraphics[width=0.24\textwidth]{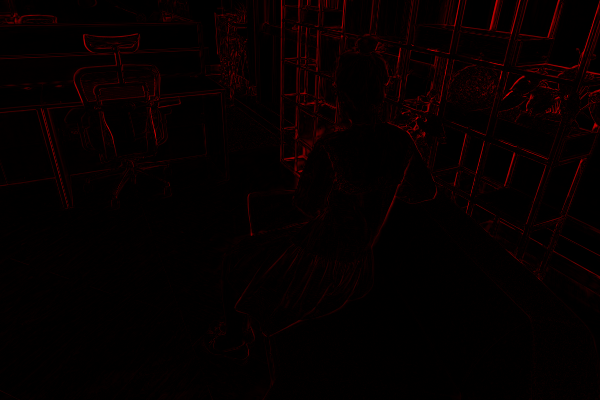} \\
			
			\includegraphics[width=0.24\textwidth]{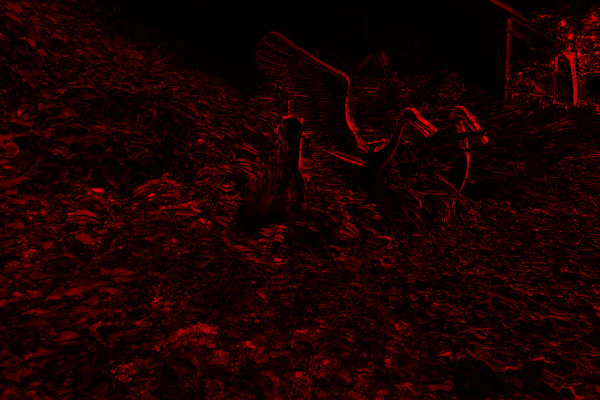} &
			\includegraphics[width=0.24\textwidth]{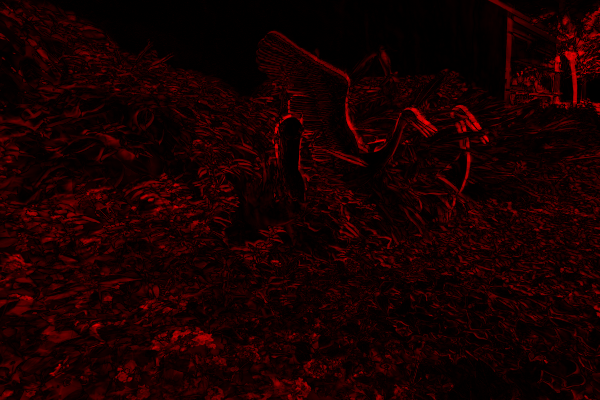} &
			\includegraphics[width=0.24\textwidth]{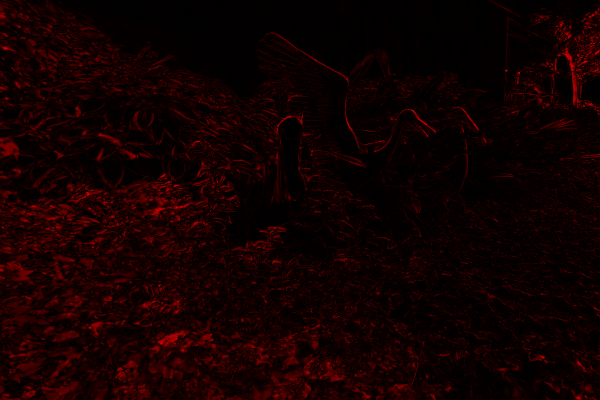} &
			\includegraphics[width=0.24\textwidth]{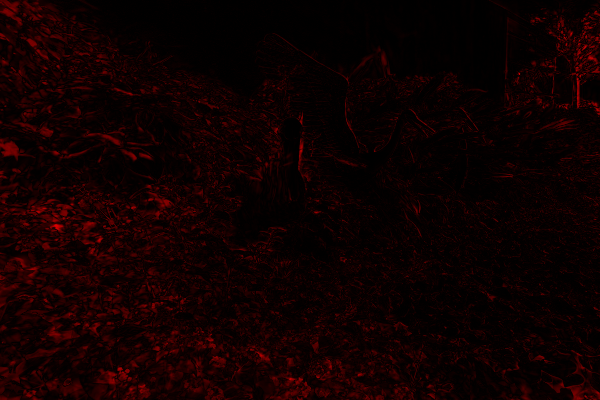} \\
			
		\end{tabular}
		\captionof{figure}{Error Map Comparison.}
		\label{fig:pose}
	\end{center}
\end{table*}
\clearpage
\clearpage
{
    \small
    \bibliographystyle{ieeenat_fullname}
    \bibliography{main}
}
